%% file: main.tex
\documentclass{article}

\usepackage{etoolbox}

\usepackage[final, nonatbib]{Styles/neurips_2024}

\usepackage[utf8]{inputenc} 
\usepackage[T1]{fontenc}    
\usepackage{hyperref}       
\usepackage{url}            
\usepackage{booktabs}       
\usepackage{amsfonts}       
\usepackage{nicefrac}       
\usepackage{microtype}      
\usepackage{xcolor}         

\usepackage{wrapfig}
\usepackage{amsmath}
\usepackage{amssymb}
\usepackage{amsthm}
\usepackage{booktabs}
\usepackage{subcaption}
\usepackage{array}
\usepackage{enumitem}
\usepackage{makecell}
\usepackage{diagbox}
\usepackage{bm}
\usepackage{dsfont}
\usepackage{multirow}
\usepackage{varwidth}
\usepackage{pifont}
\usepackage{makecell}
\usepackage{wrapfig}
\usepackage{graphicx} 
\usepackage{comment}
\usepackage{color}
\usepackage[colaction]{multicol}
\usepackage{colortbl}
\usepackage{algpseudocode} 
\usepackage{algorithm}
\usepackage{xspace}
\usepackage{rotating}
\usepackage{longtable}
\usepackage{adjustbox}
\usepackage{threeparttable}
\usepackage{listings}

\definecolor{codegreen}{rgb}{0,0.6,0}
\definecolor{codegray}{rgb}{0.5,0.5,0.5}
\definecolor{codepurple}{rgb}{0.58,0,0.82}
\definecolor{backcolour}{rgb}{0.95,0.95,0.92}

\lstdefinestyle{mystyle}{
    backgroundcolor=\color{backcolour},   
    commentstyle=\color{codegreen},
    keywordstyle=\color{magenta},
    numberstyle=\tiny\color{codegray},
    stringstyle=\color{codepurple},
    basicstyle=\ttfamily\footnotesize,
    breakatwhitespace=false,         
    breaklines=true,                 
    captionpos=false,                    
    keepspaces=true,                 
    numbersep=5pt,                  
    showspaces=false,                
    showstringspaces=false,
    showtabs=false,                  
    tabsize=2
}

\usepackage{caption}
\usepackage{minitoc}
\renewcommand \thepart{}
\renewcommand \partname{}

\definecolor{cyl_color}{rgb}{0.858, 0.188, 0.478}

\definecolor{customblue}{HTML}{007acc}
\definecolor{customred}{HTML}{cc3300}
\definecolor{customgray}{HTML}{85888d}
\definecolor{customdarkblue}{HTML}{445277}

\newcommand{\aname}{\textsc{DHA}\xspace}
\title{\aname: Learning Decoupled-Head Attention from Transformer Checkpoints via Adaptive Heads Fusion}

\author{%
Yilong Chen$^{1,2}$\thanks{~denotes equal contribution. $^\dagger$ denotes corresponding author. $^\ddagger$ denotes project lead.},
~Linhao Zhang$^{3*}$,
~Junyuan Shang$^{3\ddagger}$,
~Zhenyu Zhang$^3$,\\
~\textbf{Tingwen Liu}$^{1,2\dagger}$\textbf{,} 
~\textbf{Shuohuan Wang}$^3$\textbf{,}
~\textbf{Yu Sun}$^3$ \\
\small  \normalsize $^1$ Institute of Information Engineering, Chinese Academy of Sciences\\
\small  \normalsize $^2$ School of Cyber Security, University of Chinese Academy of Sciences\\
\small  \normalsize $^3$ Baidu Inc.\\
\small \{\texttt{chenyilong, liutingwen\}@iie.ac.cn} \\
\small \{\texttt{zhanglinhao, shangjunyuan, zhangzhenyu07, wangshuohuan, sunyu02\}@baidu.com}\\
}

\begin{document}

\maketitle
\doparttoc 
\faketableofcontents 

\begin{abstract}\label{sec.abs}

\input{abstract}

\end{abstract}

\section{Introduction}\label{sec.intro}
\input{introduction}

\section{Background}\label{sec.bg}
\input{preli}

\section{Observation}\label{sec.obs}
\input{observation}

\section{Method}\label{sec.method}
\input{proposal}

\section{Empirical Evaluation}\label{sec.exp}
\input{experiment}

\section{Related Work}\label{sec.rw}
\input{related_work}

\section{Conclusion}\label{sec.con}

In this paper, we propose an efficient attention architecture and a method for fast converting an MHA checkpoint into an efficient structure. By grouping similar heads and performing controlled linear fusion, we develop an initial DHA architecture that decouples head components at various layers, reducing training overhead while maintaining performance. Experimental results show that our method preserves the knowledge of the original model, improving training acceleration, inference efficiency, and computational cost savings. This transformation paradigm offers research value and potential for broader application with minimal performance loss and reduced computational effort.

\section*{Acknowledgments}
We would like to thank Yinqi Yang, Jiawei Sheng, Xinhua Zhang, Shicheng Wang, Chuanyu Tang and members of the IIE KDsec NLP group for their valuable feedback and discussions. 
We are very grateful to Mengzhou Xia for providing the concise and effective ShearingLLaMA experimental code and for her assistance during the reproduction process. Work done during Yilong Chen’s internship in Baidu Inc. This research is supported by the National Key Research and Development Program of China (grant No.2021YFB3100600) and the Youth Innovation Promotion Association of CAS (Grant No. 2021153).

\clearpage

\bibliographystyle{unsrt}
\bibliography{reference}

\newpage
\appendix
\addcontentsline{toc}{section}{Appendix} 
\renewcommand \thepart{} 
\renewcommand \partname{}
\newpage
\part{Appendix} 
\input{appendix}




\input{checklist}

\end{document}

%% file: abstract.tex
Large language models (LLMs) with billions of parameters demonstrate impressive performance. However, the widely used Multi-Head Attention (MHA) in LLMs incurs substantial computational and memory costs during inference. 
While some efforts have optimized attention mechanisms by pruning heads or sharing parameters among heads, these methods often lead to performance degradation or necessitate substantial continued pre-training costs to restore performance.
Based on the analysis of attention redundancy, we design a Decoupled-Head Attention (DHA) mechanism.
DHA adaptively configures group sharing for key heads and value heads across various layers, achieving a better balance between performance and efficiency.
Inspired by the observation of clustering similar heads, we propose to progressively transform the MHA checkpoint into the DHA model through linear fusion of similar head parameters step by step, retaining the parametric knowledge of the MHA checkpoint. We construct DHA models by transforming various scales of MHA checkpoints given target head budgets.
Our experiments show that DHA remarkably requires a mere 0.25\% of the original model's pre-training budgets to achieve 97.6\% of performance while saving 75\% of KV cache. Compared to Group-Query Attention (GQA), DHA achieves a 5$\times$ training acceleration, a maximum of 13.93\% performance improvement under 0.01\% pre-training budget, and 4\% relative improvement under 0.05\% pre-training budget.

%% file: introduction.tex
Transformer-based large language models (LLMs) shine in various natural language tasks due to their powerful understanding and generation capabilities~\cite{claude,Gpt-4,touvronLlamaOpenFoundation2023}. Multi-Head Attention (MHA) is widely used in LLMs, with the number of heads increasing as the model size grows. However, MHA inference overhead increases linearly with the expansion of the context and model sizes, due to the surprisingly large memory consumption of the \textit{KV Cache} mechanism. For instance, a 7 billion-parameter model
with 32 heads and 32 layers, an input batch size of 4, and a sequence length of 32k results in 64GB of KV cache, which is $4.7\times$ larger than the model weights.



To reduce computational and memory overhead during inference, a widely used approach involves adapting the MHA model to a more efficient structure through the reuse of parameters across multiple heads~\cite{shazeer2019fast, ainslie2023gqa, javadi2023gqkva}
, such as Multi-Query Attention (MQA)~\cite{shazeer2019fast} and Grouped-Query Attention (GQA)~\cite{ainslie2023gqa}. 
These methods utilize a portion of the original training computation which avoid information loss due to training-inference inconsistencies, a common issue in pruning-based~\cite{agarwal2024chai, h2o, liu2023scissorhands, ge2023model, sink} works. However, the training computation is prohibitively expensive for recovering the model's performance, due to the information loss in the parameters when creating the initial point.

Thus, in this work, we seek to address the following question:

\begin{center}
\emph{How can we construct a \textbf{more efficient} model while keeping \textbf{costs as low as possible}?} \\
\end{center}

\begin{figure}[t]
\subfloat[Head Parameter Similarity in 0th Layer]{\includegraphics[width=0.5\textwidth]{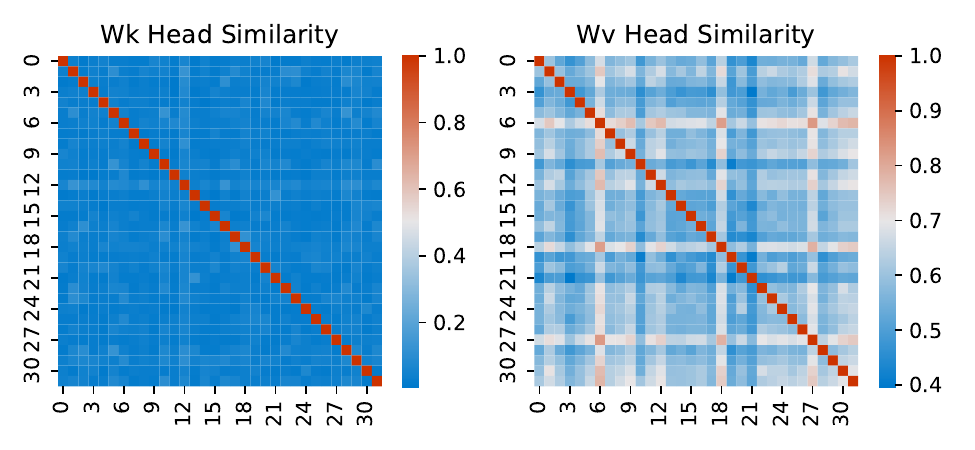}\label{fig:7b_l0}}
\subfloat[Head Parameter Similarity in 21st Layer]{\includegraphics[width=0.5\textwidth]{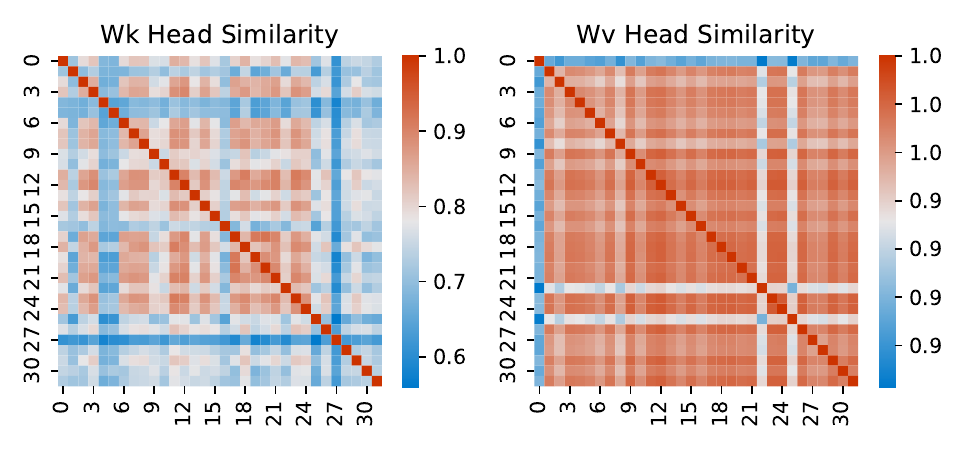}\label{fig:7b_l21}}
\caption{Visualization of the similarity between heads within the MHA of LLaMA2-7B model at the 0th layer (a) and the 21st layer (b). Details in Appendix~\ref{app:obsmha}. Key heads and value heads exhibit decoupled distributions.}
\vspace{-2mm}
\end{figure}

With the limited understanding of parameter characteristics in modern LLMs, we first perform an empirical analysis from the perspectives of heads' parameter similarity. We observe that there are some head-clusters with high internal similarity in MHA checkpoints. Similar head clusters imply a enormous redundancy in MHA, which coincides with the sparsity found in previous studies~\cite{qinExploringModeConnectivity2022,zhangCRaShClusteringRemoving2023}. 
In particular, the clusters of key heads and value heads across different layers show a \textbf{decoupled} distribution, meaning that there is a significant variation in the distribution of head-cluster similarities across layers, key heads and value heads, as illustrated in Fig.~\ref{fig:7b_l0},\ref{fig:7b_l21}.
Intuitively, we can prune redundant heads based on the above characteristics. Nonetheless, each head has its unique role, and thus no heads should be arbitrarily discarded. Furthermore, we find that linear fusion based on multiple similar heads can reconstruct the original head functionality without causing a  significant performance drop (see Sec.~\ref{obs:cluster}). Based on this observation, we believe that selectively fusing corresponding heads in clusters can construct a more efficient architecture with minimum loss.
%

\begin{figure}[t]


\centering
\includegraphics[width=\textwidth]{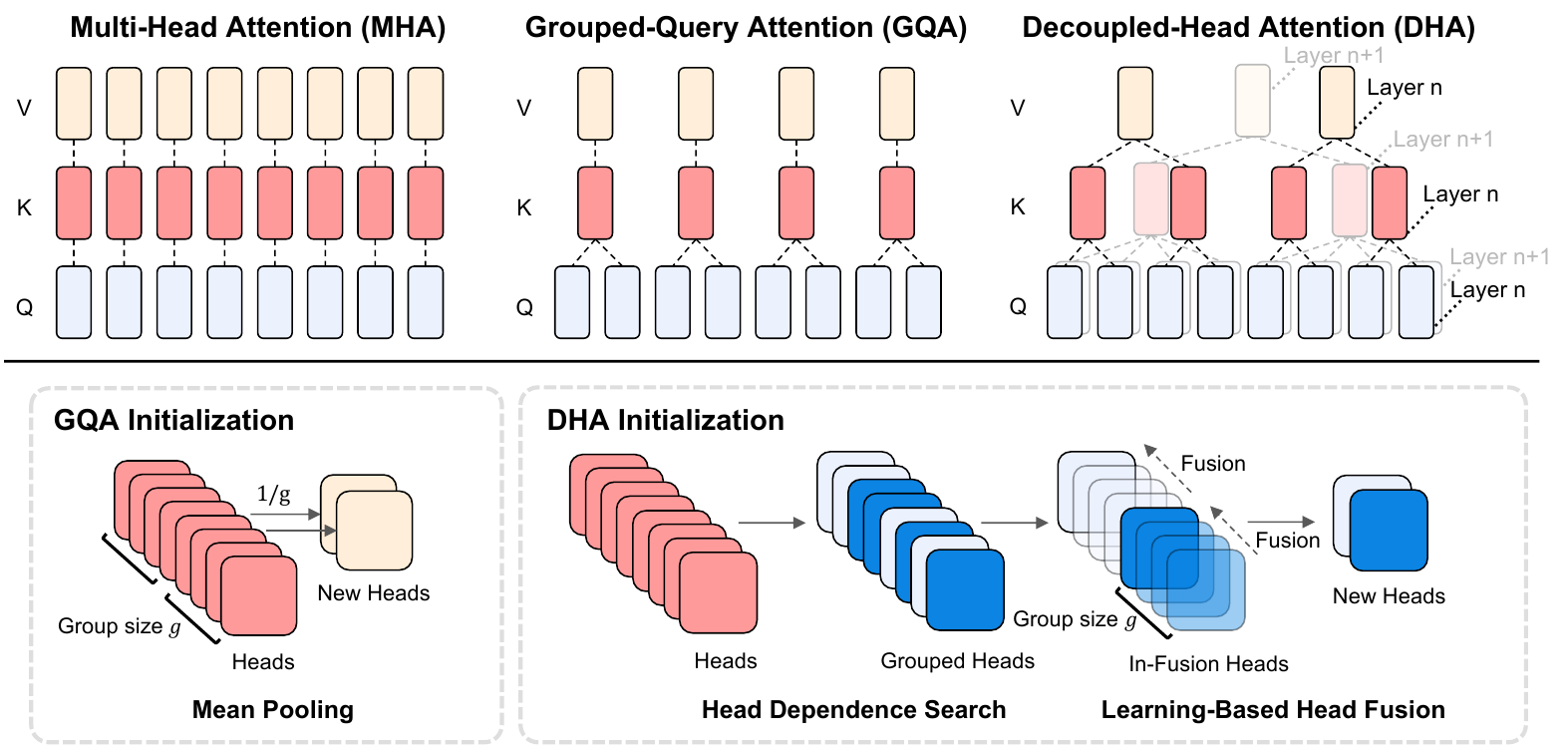}
\caption{
 \textbf{Upper: }Overview of Decoupled-head method. Multi-Head attention (MHA) has equal query, key and value heads. Grouped-Query attention (GQA) instead shares single key and value heads for each group of query heads. Decoupled-Head attention (DHA) shares key heads and value heads for different groups of query heads in different layers.
\textbf{Lower:} GQA Initialization: Heads are mean pooled into a single head; DHA Initialization: DHA search head grouping and progressively fuse heads to maintain parameter functions. 
}\label{fig:intro}
\vspace{-2mm}
\end{figure}

In this paper, we propose \textbf{Decoupled-Head Attention (\aname)}, an efficient attention architecture developed through the \textbf{Adaptive Head Fusion} of checkpoints' parameters. Recalling the decoupled heads parameter characteristics, DHA allocates different numbers of key heads and value heads at different layers to balance model efficiency and performance. The MHA checkpoint can be rapidly transformed into DHA with three stages: \textbf{Search}, \textbf{Fusion}, and \textbf{Continued Pre-training (CT)}.
During the Search stage, we group similar functional heads together and determine reasonable allocations of key heads and value heads for each layer. Specifically, we reconfigure the original key and value head into multiple linear combinations of heads within the same layer. Thus, we can allocate the heads based on the loss after replacement.
In the Fusion stage, we perform linear fusion on similar heads, ensuring the preservation of original functionality. Leveraging the Augmented Lagrangian approach~\cite{rush-etal-2010-dual,wangStructuredPruningLarge2020a}, the Fusion operator initializes from MHA and explores possible head combinations in the early training, followed by refined intra-group head fusion in the later.
Based on well-trained operators on unlabeled data, we can rapidly obtain high-performing initial points for DHA from MHA checkpoints, requiring only a minimal amount of Continued Pre-training to restore performance.


To verify the effectiveness, we construct DHA on models of different sizes, such as LLaMA2-7B~\cite{touvronLlamaOpenFoundation2023}, Sheared-LLaMA-2.7B \& -1.3B~\cite{xiaShearedLLaMAAccelerating2023} with the heads budget ratio set at 50\% and 25\%. 
 With a modest fusion training of just 0.2 billion tokens, \aname learns sufficiently competent initial points. As the continued pretraining progresses, DHA continuously outperforms GQA narrowing the gap with MHA on 9 representative downstream tasks. \aname only requires  0.25\% of MHA pre-training budget. Meanwhile, \aname is capable of reducing \textit{KV Cache} by up to 75\% compared to MHA with minimal accuracy trade-off (maximum of 5.6\%). Compared to GQA, DHA achieves a 5$\times$ training acceleration, a maximum 13.93\% performance improvement under 0.01\% pre-training budget, and 4\% relative improvement under 0.05\% pre-training budget. 
Overall, \aname exhibits great performance and efficiency, which can be quickly adapted to various existing MHA Transformer models.\vspace{-2mm}

%% file: preli.tex
Let $\mathbf{X}=\left(\mathbf{x}_1, \ldots, \mathbf{x}_p\right) \in \mathbb{R}^{p \times d_{\text{model}}}$ denote the input prompts of hidden states of a Transformer layer, where $p$ stands for the number of tokens and $d_\text{model}$ for the hidden state dimension. 

\paragraph{Multi-Head Attention (MHA)} MHA~\cite{vaswani2017attention} performs the attention with $H$ different heads. For $h$-th head, different weight matrices $\textbf{\rm W}_q^h, \textbf{\rm W}_k^h, \textbf{\rm W}_v^h \in \mathbb{R}^{d_{\text{model}} \times d_k}$ are used to project the input sequence into query, key, value vector, 
where $d_k$ represents head dim. Denote softmax funcion as $\sigma$, we have:
\begin{equation}
\small
\text {MHA}=\operatorname{Concat}\left(\text {head}_1, \ldots, \text{head}_{\text {H }}\right) \textbf{\rm W}_O, \\
\text {where } 
\text{head}_{h} = \sigma\left(\mathbf{X}\textbf{\rm W}_q^h (\mathbf{X}\textbf{\rm W}_k^h)^T \cdot \frac{1}{\sqrt{d_k}}\right) \mathbf{X}\textbf{\rm W}_v^h
\end{equation}
Ultimately MHA combines heads' outputs through the output projection $\mathbf{\rm W}_O \in \mathbb{R}^{d_\text{model} \times d_\text{model}}$\footnote{MHA consists of $H$ heads and stores a $2 \times H \times p \times d_{\text{model}}$ dimension KV cache for accelerating inference.}.

\paragraph{Grouped-Query Attention (GQA) \& Multi-Query Attention (MQA)} To accelerate inference, MQA~\cite{shazeer2019fast} and GQA~\cite{ainslie2023gqa} have been proposed based on the idea of reusing head parameter weights. In these variants, $H$ different query heads are divided into $G$ groups, where the heads within the same group share the same key heads and value heads parameter matrices.
Given the mapping relationship from the $h$-th query head to a GQA key and value heads using the many-to-one function $g(h)$, we define the $h$-th head forward pass as:
\begin{equation}
\small
\text{head}_{h}=\sigma\left({\mathbf{X}\textbf{\rm W}_q^h(\mathbf{X}\textbf{\rm W}_k^{g(h)}})^T\frac{1}{\sqrt{d_k}}\right)\mathbf{X}\textbf{\rm W}_v^{g(h)},
\text {where } 
\textbf{\rm W}_{k/v}^{{g(h)}_{\text{init}}}  
=\frac{\sum_{
\textbf{\rm W}_{k/v}\in {\mathbb{K}/\mathbb{V}}_{{g(h)}_{\text{init}}}
}\textbf{\rm W}_{k/v}}{|{\mathbb{K}/\mathbb{V}}_{{g(h)}_{\text{init}}}|}
\label{eq:gqa}
\end{equation}
Here, ${\mathbb{K}/\mathbb{V}}_{{g(h)}_{\text{init}}}$ refers to MHA key/value heads parameters within the ${g(h)}_{\text{init}}$-th group during GQA initialization. When transitioning from an MHA checkpoint, GQA uses the mean pooling method for heads within the group. MQA is a special case of GQA where $G = 1$\footnote{GQA and MQA consist of $H + 2 \times G$ heads in total and store a $2 \times G \times p \times d_{\text{model}}$ dimension KV cache.}. 

 

Due to mean pooling for initialization, GQA results in loss of capability when converting from MHA, necessitating expensive pre-training to recover. We aim to identify better initialization and more refined head mapping relationships to achieve superior performance with reduced training costs.

%% file: observation.tex

To study the inherent characteristics of head parameters in MHA, we use Centered Kernel Alignment~\cite{kornblith2019similarity} to calculate the heads' similarity within each layer's $\textbf{\rm W}_k, \textbf{\rm W}_v$. Based on the average heads' similarity, we define the redundancy of each MHA layer. For details, please refer to Appendix~\ref{app:cka}.

\subsection{Head clusters in MHA}\label{obs:cluster}
\paragraph{Observation} From Fig.~\ref{fig:7b_l0} and Fig.~\ref{fig:7b_l21}, we observe that clusters form spontaneously among heads, with high similarity within clusters and low similarity between clusters. 
It indicates that heads among different clusters may have distinct functionalities, processing linguistic features in various aspects.

\paragraph{Analysis}Given the numerous similar head clusters in $\textbf{\rm W}_k$ and $\textbf{\rm W}_v$, we identified the opportunity to linearly fuse functionally similar heads within clusters while retaining each head's parameterized knowledge. 
We conducted an empirical study, transforming the parameters of Head 0 in MHA into a linear fusion of the parameters from Heads 0, 1, 2, and 3. We share the fusion head across four query heads and progressively optimize the fusion ratio under the LmLoss. For details, please refer to Sec.~\ref{sec:fusion}. As shown in Fig.~\ref{fig:loss_ratio}, the loss remains unchanged as the proportion of Head 0 decreases and only increases when four heads parameters' ratios approach an even distribution. It suggests that fusing similar parameters can reduce the number of heads without significant information loss.

\subsection{Variability across Layers and KV pairs} 
\begin{wrapfigure}[12]{r}{0.57\textwidth}
\vspace{-0.4cm}
  \centering
  \subfloat[Loss in 4-Head Fusion]{\includegraphics[width=0.305\textwidth]{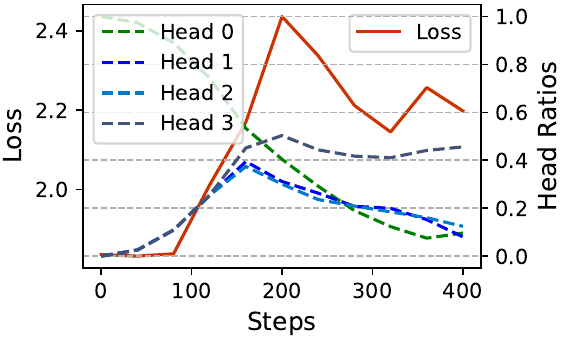}\label{fig:loss_ratio}} \hfill
  \subfloat[32-Layer Redundancy]{\includegraphics[width=0.26\textwidth]{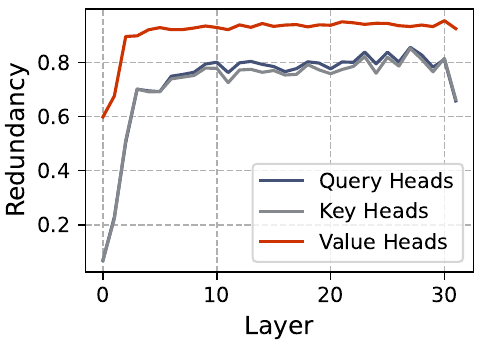}\label{fig:redundancy}}
  \caption{(a) Model loss with heads proportions in linear fusion. (b) Layer Redundancy of the query, key, value head parameter matrices in the LLaMA2-7B model MHA.}
  \vspace{-0.5cm}
\end{wrapfigure}
\paragraph{Observation} The distribution of similar head clusters varies between different layers. As illustrated in Fig.\ref{fig:7b_l0},~\ref{fig:7b_l21}, the 0th layer of MHA shows few similar head clusters, while the 21st layer exhibits many. Within the same layer, value heads exhibit more clusters and higher similarity compared to key heads, indicating a divergence between the two. 
Fig.~\ref{fig:redundancy} shows that the redundancy is lower in the initial and final layers, and higher in the middle layers. Moreover, $\textbf{\rm W}_v$ redundancy significantly exceeding that of $\textbf{\rm W}_k$.

\paragraph{Analysis}
Inspired by layer and key-value head variability, we propose allocating more heads to layers with lower redundancy to enhance learning and expression. Since $\textbf{\rm W}_v$ shows higher redundancy than $\textbf{\rm W}_k$, 
we can decouple and allocate more heads budget to critical key components, while compressing redundant value heads at a higher compression rate. Finer grouping and sharing based on the parameters function may contribute to compression rates and performance improvements.

%% file: proposal.tex
In this section, we propose a more efficient Decoupled Head Attention (\aname) architecture and its construction process. We define DHA in Sec.~\ref{sec:dhadef} and Adaptive Head Fusion algorithm in Sec.~\ref{sec:fusion}. Then we demonstrate the adaptive construction based on the MHA checkpoint, which can be divided into: \textbf{Search}, \textbf{Fusion}, and \textbf{Continued Pre-training} (Discussed in in Sec.~\ref{sec:search}). 
Finally, we introduce practical application of our \aname architecture on the LLaMA2 model in Sec.~\ref{sec:search}.
\subsection{Decoupled-Head Attention (\aname)} \label{sec:dhadef}

We present a more efficient attention architecture called Decoupled-Head Attention (DHA). Based on observed significant functional differences among different layers' key value heads, \aname adaptively allocates more heads to critical components, thus enhancing overall model efficiency and performance.

\paragraph{Definition} Defined model with $L$ layers and $H^{\text{Q}}$ heads in a layer, the numbers of Key heads and Value heads in the \( l \)-th layer are denoted as $H_l^{\text{K}},H_l^{\text{V}}$.
We define the \textbf{many-to-one} mapping functions $d^{\text{K}}(h,l)$ and $d^{\text{V}}(h,l)$ representing key and value head corresponding to the $h$-th query head in $l$-th DHA layer. 
The computation be formalized as follows:
\begin{equation}
\small
\text{head}_{h,l} = \sigma\left(\mathbf{X}\textbf{\rm W}_q^h (\mathbf{X}\textbf{\rm W}^{d^{\text{K}}(h,l)}_k)^T \cdot \frac{1}{\sqrt{d_k}}\right)\mathbf{X}\textbf{\rm W}^{d^{\text{V}}(h,l)}_v
\label{eq:dha}
\end{equation}

DHA shares a key and value head in multi query heads' computation based on independent mapping functions at different layers\footnote{\aname consists of $H = H^{\text{Q}} + \sum_{l=1}^{L} H_l^{\text{K}} + \sum_{l=1}^{L} H_l^{\text{V}}$ heads in total. }. GQA can be considered a special case of DHA, where not only all layers share the same mapping functions, but the mapping functions for keys and values are identical.


\subsection{Learning Efficient MHA Transformation via Linear Heads Fusion} \label{sec:fusion}
\begin{figure}[t]


\begin{minipage}[c]{\textwidth}
\centering
  \includegraphics[width=\textwidth]{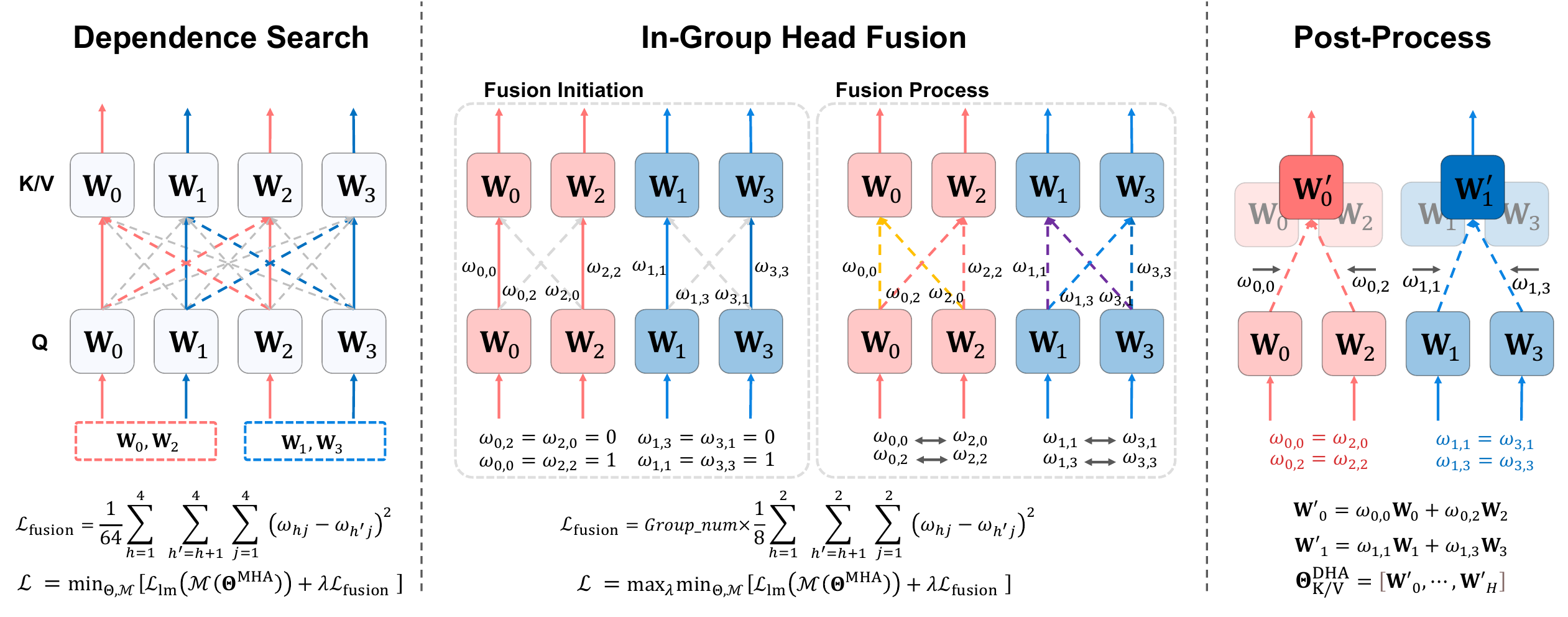}
  \caption{An illustration of \aname. First, we reconstruct the  a single head forward as a linear combination of multiple heads' forward with proportions $\mathbf{\omega}$, grouping heads with similar functions based on multi-step optimization. Next, we initialize and optimize the fusion operators. $\Leftrightarrow$ indicates the optimization narrows the distance between proportions $\mathbf{\omega}$. Finally, we fuse heads within groups and continued pre-training DHA model.}\label{fig:DHA}
\end{minipage}
\vspace{-2mm}
\end{figure}

Due to the high cost of building an efficient Attention mechanism in LLM from scratch, we construct DHA based on the existing MHA checkpoint using minimal computational budgets.
Based on the head clustering phenomenon in MHA, we propose a linear fusion method for similar heads within clusters. By incrementally fusing head parameters, we compress the number of heads while retaining the original model's knowledge, significantly reducing training budgets and improving performance.

\paragraph{Goal}Formally, we define a model with Layer number $L$ and Head number $H$ as $\boldsymbol{\Theta}_{L,H}=\left[\textbf{\rm W}_1, \cdots, \textbf{\rm W}_L\right]$, where $\textbf{\rm W}_l \in \mathbb{R}^{D \times D}$ denotes the weight of layer $l$ with input and output dimension $D$. In the initialization, our goal is to transfer knowledge from a MHA model $\boldsymbol{\Theta}_{{L}, H_{1}}^{\text{MHA}}$ to a DHA model $\boldsymbol{\Theta}_{{L}, H_{2}}^{\text{DHA}}$, where $H_1 > H_2$ .
By learning a fusion operation that minimizes the functional difference between MHA and DHA model, the goal can be formalized as
\begin{equation}
\small
\begin{aligned}
\underset{\boldsymbol{\Theta},\mathcal{M}}{\arg \min } 
&~\mathbb{E}_{\boldsymbol{x} \sim \mathcal{D}}~
\left[
\mathcal{L}_{\text{lm}}\left(\boldsymbol{x} ; \mathcal{M}(\boldsymbol{\Theta}^{\text{MHA}})\right)
+
\lambda\mathcal{L}_{\text{fusion}}\left(\boldsymbol{x} ; \mathcal{M}(\boldsymbol{\Theta}^{\text{MHA}}), \boldsymbol{\Theta}^{\text{DHA}}\right)
\right]
\end{aligned}
\label{eq:1}
\end{equation}
Where $\mathcal{M}$ is the fusion operator, $\mathcal{D}$ is the training dataset, $\mathcal{L}_{\text{lm}}$ is the training loss function, $\mathcal{L}_{\text{fusion}}$ measures the transformation from MHA to DHA, and $\lambda$ is the learnable scale factor.

\paragraph{Fusion Operator}During DHA initialization, the fusion operator $\mathcal{M}$ constructs new heads based on the linear combinations of the original key and value heads within the group, and shares the new heads among the query heads' forward. Define each group $\mathbb{K}_{d^{\text{K}}(h,l)},\mathbb{V}_{d^{\text{V}}(h,l)}$ represents key, value heads group corresponding to the $h$-th query head in $l$-th layer, $g=\left\{g^{\text{K}},g^{\text{V}}\right\}$ as the group size. 
By introducing variables $\mathbf{\omega}_h=\left\{\mathbf{\omega}_{hj}\right\}_{j=1}^{g},\mathbf{\omega}\in\mathcal{M}$ 
represents the proportion of $j$-th key, value head involved in the $h$-th query head forward within group. For each group, a head have forward pass as:
\begin{equation}
\small
\text{head}_{h,l} = \sigma\left(\mathbf{X}\textbf{\rm W}_q^h (\mathbf{X}\textbf{\rm W}^{d^{\text{K}}(h,l)}_k)^T \cdot \frac{1}{\sqrt{d_k}}\right)\mathbf{X}\textbf{\rm W}^{d^{\text{V}}(h,l)}_v
,\text{where } 
\textbf{\rm W}_{k/v}^{d^{\text{K/V}}(h,l)}=\sum_{j=1}^{g^{\text{K/V}}} \mathbf{\omega}_{hj} \textbf{\rm W}_{k/v}^{j}
\label{eq:dhainit}
\end{equation}
where $\mathbf{\omega}_{hj}$ will be initialized to Kronecker delta function, which equals 1 if and only if $h = j$, and equals 0 otherwise. Under this initialization setting, the forward computation of DHA is completely equivalent to that of MHA, see Fig.~\ref{fig:DHA}.
\paragraph{Optimization} 
During the optimization phase, we design a fusion loss to optimize the initialized model towards DHA target architecture. 
Note that after initialization, the mapping of heads within the group $\textbf{\rm W}_q^{h,l} \rightarrow \textbf{\rm W}_{k/v}^{j}$ is a \textbf{many-to-many} mapping, denoted by the function $d^{\text{K/V}}_\text{init}(h,l)$. 
This indicates that in the forward process of each query, the key head or value head can be expressed as different linear combinations of $g$ MHA heads. According to Eq.~\ref{eq:dha}, we aim to achieve a \textbf{many-to-one} mapping that a single fused key head or value head are shared across multiple query heads in DHA, denoted by the function $d^{\text{K/V}}(h,l)$. Thus, we design a fusion loss $\mathcal{L}_{\text{fusion}}$ to optimize the initial mapping functions to converge to a single mapping function, i.e., $d_\text{init}^{\text{K/V}}(h,l) \rightarrow d^{\text{K/V}}(h,l)  , \forall h \in \mathbb{K}_n/\mathbb{V}_n$.
Specifically, we define the optimization objective as minimizing the difference between the mapping functions of different query heads $h$ and $h'$ within the $l$-th layer and $n$-th group:

\begin{equation}
\small
\mathcal{L}_{\text{head}_{l}^n}(h, h') = \frac{1}{g} \left\| \sum_{j=1}^g \omega_{hj} \textbf{\rm W}_{k/v}^{j} - \sum_{j=1}^g \omega_{h'j} \textbf{\rm W}_{k/v}^{j} \right\|^2 
= \frac{1}{g} \left( \sum_{j=1}^g (\omega_{hj} - \omega_{h'j}) \textbf{\rm W}_{k/v,ij}^{j} \right)^2
\end{equation} 
where $g=g^{\text{K/V}}$ represents the number of heads within a group. Since $W_{k/v,ij}^{j,\text{MHA}}$ can be regarded as an orthogonal scalar, and thus we only need to optimize fusion variables $\omega$ , so we have:
\begin{equation}
\small
    \mathcal{L}_{\text{fusion}} =  \sum_{l=1}^{L}\sum_{n=1}^{N}\sum_{h=1}^{g}\sum_{h'=h+1}^{g}\mathcal{L}_{\text{head}_{l}^n}(h, h'), 
    \text{subject to } 
    \mathcal{L}_{\text{head}_{l}^n}(h, h') 
= \frac{1}{g} \sum_{h=1}^g \sum_{j=1}^g (\omega_{hj} - \omega_{h'j})^2
\label{eq:dhaloss}
\end{equation}Where \( N \) represents the number of groups, \( N = \frac{H_1}{g} \). The fusion loss can be measured as the mean squared error loss of the head and head fusion variables within each group at each layer.
\paragraph{Augmented Lagrangian approach}When the fusion loss is zero, the key and value heads corresponding to query heads within the group are optimized to share the same fusion variables. This allows the new DHA key-value head parameters to be effectively shared among the queries in the group. Given that it is challenging to optimize the loss to a very small value, we use an augmented Lagrangian approach~\cite{rush-etal-2010-dual,wangStructuredPruningLarge2020a} for incremental architectural transformations. Define $t$ as the target loss,  $b$ as the base decay factor, $s$ as the current global step, $k$ as the total number of steps in the warm-up phase, the overall training optimization is an adversarial game:
\begin{equation}
\small
    \max_{\lambda} \min_{\boldsymbol{\Theta},\mathcal{M}} \mathbb{E}_{\boldsymbol{x} \sim \mathcal{D}}
    \left[
    \mathcal{L}_{\text{lm}}\left(\boldsymbol{x} ; \mathcal{M}(\boldsymbol{\Theta}^{\text{MHA}})\right)
    +
    \lambda\operatorname{max}\left(
    \mathcal{L}_{\text{fusion}} - t, 0\right)
    \right]
    ,\text{where } 
    t = \max\left(0, b^{s}\left(1 - \frac{s}{k}\right)\right)
    \label{eq:final_loss}
\end{equation}Our Augmented Lagrangian approach enforces the constraint $\mathcal{L}_{\text{fusion}} \leq t$, where the Lagrange multiplier $\lambda$ is updated during training. The update increases the loss unless the constraint is satisfied. Early in training, the model tolerates more significant discrepancies between head weights, promoting exploration. As training progresses, the margin shrinks, enforcing stricter adherence to minimizing discrepancies and refining head alignment within the group.

\subsection{Adaptive DHA Transformation on LLaMA Model} \label{sec:search}
Based on the observation of similar head clusters and key-value head parameter variability across layers, DHA employs the adaptive transformation. It allows DHA to search for and fuse similar heads while allocating different group sizes across layers. As shown in Fig.~\ref{fig:DHA}, the transformation can be divided into three stages: \textbf{Search}, \textbf{Fusion} and \textbf{Continued Pre-training}.

In the beginning, we initialize the DHA operators to the MHA model. Next, we perform 240 \textbf{Search} steps, calculating $\mathcal{L}_{\text{fusion}}$ for each layer and $\mathcal{L}_{\text{head}}$ for all heads. Based on the $\mathcal{L}_{\text{head}}$, we perform head grouping intending to minimize the average loss of heads within each group and maximize the average loss of heads between groups and groups. Based on $\mathcal{L}_{\text{fusion}}$, we use a dynamic programming algorithm to allocate more head budget to layers with higher loss within a total budget. It allows us to fuse the most similar heads to minimize loss during the fusion process and selectively compress the model's most redundant components. For more details, see Apendix~\ref{app:cluster},~\ref{app:layerallo}.

During \textbf{Fusion} phase, we modified the forward propagation path of MHA in the form of DHA based on the layer head budget and head grouping obtained during the \textbf{Search} phase. Then  we antagonistically optimize the fusion operator and update Lagrangian multipliers $\lambda$, the $\mathcal{L}_{\text{fusion}}$ that marks this DHA fusion process decreases.  When $\mathcal{L}_{\text{fusion}}$ is less than 1e-3, we terminate the fusion algorithm and enter the \textbf{Continued Pre-training} phase.

During the \textbf{Continued Pre-training} phase,  we fuse MHA head parameters based on averaged fusion weights to construct DHA initialization. DHA initialization  can recover the performance  with a small amount of restorative pre-training. For more information, please refer to Appendix~\ref{app:implement}.

Our method can theoretically transform MHA architecture in any transformer model to efficient DHA architecture. Using LLaMA models as case studies, we implemented DHA transformation with various compression rates on all MHA layers. Notably, we expanded the dimension of each head's fusion coefficient \(\omega\) from 1 to the head's dimension \(d_k\), allowing for finer-grained parameter fusion and better knowledge retention. Intuitively, we learn different fusion ratios for each dimension of the head. Only a very small number of additional parameters need to be introduced,  DHA significantly accelerates training and improves performance.

%% file: experiment.tex
\subsection{Experimental Setup} \label{sec:exp.setup}

\paragraph{Data.} To train~\aname operators and extend pre-training, we employ the RedPajama~\cite{Redpajama}, which parallels the LLaMA training data across seven domains: CommonCrawl, C4, GitHub, Wikipedia, Books, ArXiv, and Stack-Exchange. This dataset comprises a validation set with 2 million tokens, a training set containing 4 billion tokens and an additional pre-training set totaling 50 billion tokens.

\paragraph{Training.} Our experimental framework utilizes the Sheared-LLaMA codebase \cite{xiaShearedLLaMAAccelerating2023} implemented on the Composer package \cite{mosaicml2022composer}, and is executed on 8 NVIDIA A100 GPUs (80GB). The models are trained with a sequence length of 4096, employing a global batch size of 64 during the fusion phase and 256 during the continued pre-training phases.
 The learning rates were set at 1e-4 for language modeling loss, and 1e-2 for Lagrangian multipliers and fusion operators respectively.

\paragraph{Budget.} \aname models were trained for 1000 steps (0.2B token budget) during the fusion phases. For the continued pre-training, we trained both baseline models and \aname for up to 50000 steps (50B token budget). To evaluate the training acceleration capability of DHA, we evaluate its performance under two budget scenarios. First, we set a budget of \textbf{1B tokens} to compare the early-stage rapid convergence capabilities of \aname and GQA. Then, we set a budget of \textbf{50B tokens} to further assess the performance of \aname over a more extended training period.

\paragraph{Evaluation.} We employed the lm-evaluation-harness \cite{eval-harness} to evaluate our models. For common sense and reading comprehension tasks, we report 0-shot accuracy results for SciQ \cite{sciqa}, PIQA \cite{piqa}, WinoGrande (Wino.) \cite{WinoGrande:conf/aaai/SakaguchiBBC20}, ARC Easy(ARC-E.) \cite{clark2018think}, and HellaSwag (HellaS.) \cite{HellaSwag:conf/acl/ZellersHBFC19}, alongside 25-shot accuracy for ARC Challenge (ARC-C.) \cite{arcChallenge:journals/corr/abs-1803-05457}. In the assessments of continued QA and text understanding, we report 0-shot accuracy for LogiQA \cite{liu2020logiqa}, 32-shot BoolQ \cite{clark2019boolq}, and 0-shot LAMBADA \cite{paperno2016lambada}. All reported results were calculated with the mean and stderr of multiple experiments.

\paragraph{Instruction tuning evaluation.} To assess our models' capabilities after instruct tuning~\cite{ouyang2022instructgpt,alpaca}, we fine-tune both \aname and baseline models on 10,000 instruction-response pairs from the ShareGPT dataset~\footnote{\url{https://sharegpt.com}} and evaluate on another 1,000 instructions, using GPT-4 for response evaluator~\cite{dubois2023alpacafarm}. The win rate of our model relative to the baseline is reported. For detailed information, refer to Appendix~\ref{app:sft}.

\paragraph{Baselines.} We selected the LLaMA2-7B model and Sheared-LLaMA-2.7B\&1.3B (S.-LLaMA-2.7B\&1.3B) as the MHA baselines. For each scaled model's checkpoint, we constructed 25\% and 50\% compressed GQA and DHA models in 0.5B \& 1B tokens (0.01\%  \& 0.05\% of pretrain budget~\footnote{LLaMA2 was pre-trained on 2T data; Sheared-LLaMA pruned LLaMA on 50B RedPajama data.}).

\input{Tables/main_table}

\input{Tables/ablation}
\subsection{Experimental Results}

\paragraph{Foundational Capabilities.} 

Tab. \ref{tab:main_result} shows the foundational capabilities of DHA and GQA models at 50\% and 25\% compression rates (e.g., 64 key value heads compress to 16) across different scales. DHA was obtained by transforming LLaMA using adaptive head fusion and then further pre-trained with 1B tokens. For comparison, we constructed GQA with the same compression rates and training budget. Experiments show that DHA can achieve efficient architecture with only 0.05\% of the original model's pre-training cost without significant performance loss. Compared to GQA, DHA consistently achieved better performance across all model scales and pre-training cost settings. Under the same checkpoint and training budget settings, DHA demonstrates significant improvements at higher compression rates. For example, with LLaMA7B at a 25\% compression rate, DHA achieved a 4\% relative performance improvement over GQA. This showcases DHA's fusion algorithm's ability to efficiently retain knowledge at high compression rates and the advantage of DHA's decoupled architecture in adaptively compressing redundant components. Possibly due to the lack of relevant data, DHA performed on par with LogiQA. As shown in Fig.~\ref{fig:acc_budget}, DHA's performance advantage becomes more remarkable with reduced training budgets. It indicates that DHA effectively retains knowledge of larger models, significantly reducing pre-training costs.



\paragraph{Better Initialization.} We examined whether DHA offers a better initialization point than GQA by pre-training both DHA and GQA models on the original RedPajama dataset. Fig.~\ref{fig:loss_budget} shows that the initial loss of the GQA model is high and decreases slowly. In contrast, the DHA model starting from MHA exhibits a minor increase in LM loss as fusion progresses, maintaining a consistently lower loss. DHA converges with just 0.1B data, demonstrating a 5$\times$ training speedup compared to GQA. Fig.~\ref{fig:acc_budget} reports the average downstream task accuracy of \aname and GQA during continued pre-training. \aname achieves comparable performance to GQA's at 1B tokens with only 0.2B tokens, outperforming GQA's 0.2B token performance by 13.93\%. This demonstrates \aname's effectiveness in retaining parameter information. Ultimately, DHA achieves a higher performance ceiling than GQA due to retaining information from the original model and its more efficient architecture, whereas GQA loses information during initialization.

\subsection{Analysis} 

\begin{figure}[t]
    \centering
    \begin{minipage}[b]{0.31\linewidth}
        \centering
        \vspace{0pt}
        \includegraphics[width=\textwidth]{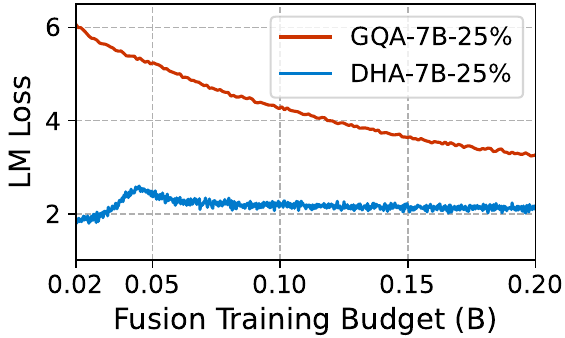}
        \caption{LM Loss with Fusion Training (B) between GQA-7B-25\% and DHA-7B-25\%. }
        \label{fig:loss_budget}
    \end{minipage}
    \hfill
    \begin{minipage}[b]{0.32\textwidth}
        \centering
    \vspace{0pt}
    \includegraphics[width=\textwidth]{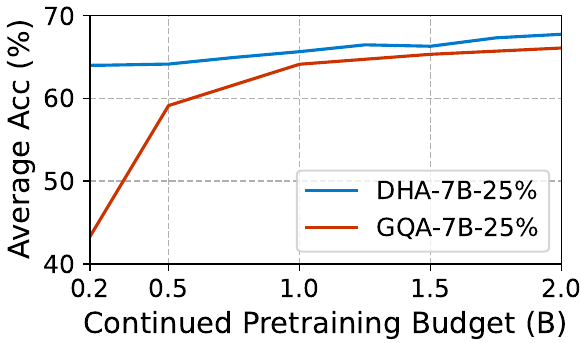}
    \caption{Task Average Accuracy (\%) with CT (B) of \aname-7B-25\% and GQA-7B-25\%. }
        \label{fig:acc_budget}
    \end{minipage}
    \hfill
    \begin{minipage}[b]{0.3\linewidth}
        \centering
        \vspace{0pt}
        \includegraphics[width=\textwidth]{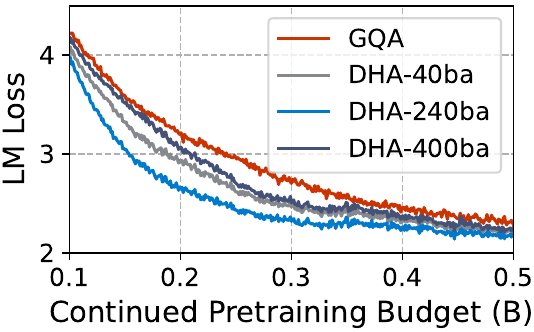}
        \caption{LM Loss with CT (B) between GQA-7B-25\% and Cold-Start DHA after X Step Search.}
        \label{fig:kv_cache}
    \end{minipage}
\end{figure}
\input{Figures/visual}

\vspace{-1mm}\paragraph{Ablation Study.} We report the effects of ablating Linear Heads Fusion and Adaptive Transformation
in Tab.~\ref{tab:ablation}. When training with less data (0.5B), ablating Linear Heads Fusion leads to significant performance degradation, indicating that this method preserves crucial knowledge in LLMs, greatly accelerates DHA model training, and enhances performance. Adaptive Transformation allocates parameters more efficiently during construction, thereby strengthening the model's capability and reducing training difficulty. When we allocate more training budget to 1B, DHA's efficient architecture after Adaptive Transformation plays a more significant role, enhancing the model's performance ceiling. When continuing to pre-train the DHA model, it demonstrated strong learning capabilities and sustained performance improvements, ultimately achieving 97.6\% of the performance with just 0.25\% of the original model's pre-training budget, while saving 75\% of the KV cache.

\vspace{-1mm}\paragraph{Training Budget Allocation.} \input{Tables/fusion_pt}Allocating more computation to the fusion phase aids in better retention of information within the checkpoint. Our experiments assessed the effects of budget allocations between the fusion and CT phases within \textbf{a fixed budget of 2 billion tokens}. Tab.~\ref{tab:pruning_finetuning} shows that increasing the fusion budget consistently from 0.05B to 0.2B improves model performance at the initialization point. Training with just 0.1B data is sufficient to achieve a good starting point, and increasing fusion budget will not affect the final performance. This experiment also demonstrates the necessity and effectiveness of the fusion stage under low-resource conditions. When we have a larger training budget, we can allocate more resources to the fusion stage to achieve a better initialization point for \aname.


\paragraph{Heads Budget Allocation.} We investigated how the model adaptively allocates decoupled head group sizes across different layers under global head budgets. As illustrated in Fig.~\ref{fig:dha_budget}, the head numbers of DHA layers decrease from higher to lower across layers. Deeper layers exhibit higher compression rates due to greater redundancy. However, the initial and crucial layers need more heads, suggesting they may have specialized functions.  As shown in Fig.~\ref{fig:kv_cache}, we presented the LM loss for the cold-start training of DHA models initialized with parameter averaging under different DHA configurations obtained at various search steps.  Despite using the same initialization method as GQA, DHA exhibits a faster loss decline and a lower final loss. This indicates that DHA's architecture can accelerate training and achieve better performance, even without Linear Heads Fusion method.

\paragraph{Parameter Characteristics in DHA.} 
For interpretability analysis, we visualized the parameter characteristics of the post-fusion DHA model in Fig.~\ref{fig:prepost} (detials in Appendix~\ref{app:obsdha}),  and compared them with those prior to fusion. The DHA parameter distribution shows consistency with MHA's. This indicates that DHA effectively aggregates multiple similar functional heads within clusters and new fused heads successfully reconstruct the functionalities of multiple origin heads in MHA. It is noteworthy that the significant reduction in the number of similar heads within the DHA architecture indicates that our method effectively reduces redundancy among the heads.

%% file: Tables/main_table.tex
\begin{table}[t]
    \caption{ Comprehensive assessment of model's fundamental capabilities, in which \aname models demonstrate competitive performance while requiring significantly fewer training resources. Models with $^\dagger$ use MHA. }
    \resizebox{1\textwidth}{!}{%
    \setlength{\tabcolsep}{2pt}
    \begin{tabular}{lccccccccccc} 
    \toprule
                              && \multicolumn{6}{c}{\textbf{Commonsense \& Comprehension}} & \multicolumn{2}{c}{\textbf{Continued}}& \textbf{LM} &                                                                                                                                  \\ 
    \cmidrule(lr){3-8}    \cmidrule(lr){9-10}   \cmidrule(lr){11-11}                           
    \textbf{Model} & \textbf{Budget} & \textbf{SciQ}          & \textbf{PIQA}             & \textbf{Wino.}& \textbf{ARC-E}              & \textbf{ARC-C}& \textbf{HellaS.}& \textbf{LogiQA}& \textbf{BoolQ}& \textbf{LAMB.}&\textbf{Average}\\
    \midrule
     LLaMA2-7B$^\dagger$ &2T& 94.1 & 78.1& 69.1& 76.3&49.7& 58.9& 25.7& 80.8& 74.1&67.4\\
     \aname-7B-50\%& 50B& 93.4& 78.5& 69.1& 73.8& 45.9& 58.6& 22.5& 79.1& 71.1&65.8\\
     \aname-7B-25\% & 50B& 92.4& 78.5& 68.6& 72.9& 43.9& 57.6& 22.4& 76.7& 70.2&64.8\\ 
    \midrule
    \rowcolor{customgray!10} GQA-7B-50\% &1B& 90.7& \textbf{76.8}& 66.5& \textbf{71.3}& 41.9& 53.6& \textbf{22.4}& 70.5& 67.0&62.3\\
    \rowcolor{customgray!10} \aname-7B-50\%&1B& \textbf{90.8}& 76.5& \textbf{66.7}& \textbf{71.3}& \textbf{44.6}& \textbf{55.1}& \textbf{22.4}& \textbf{74.8}& \textbf{67.2}&\textbf{63.3}\\
     \rowcolor{customgray!20} GQA-7B-25\%  &1B& 86.5& 74.3& 59.1& 67.6& 37.5& 49.2& \textbf{24.1}& 65.8& 58.3&58.0\\
     \rowcolor{customgray!20} \aname-7B-25\% &1B& \textbf{90.0}& \textbf{75.2}& \textbf{63.8}& \textbf{70.4}& \textbf{39.3}& \textbf{52.2}& 21.1& \textbf{72.3}& \textbf{62.9}&\textbf{60.7}\\
    \midrule
    \midrule
     S.-LLaMA-2.7B$^\dagger$ &2T& 91.2& 76.1& 64.9& 67.3& 38.8& 52.2& 22.1& 74.4& 68.3&61.7\\ 
    \rowcolor{customgray!10}  GQA-2.7B-50\%&1B& 86.7& 74.8& 59.0& 64.0& 34.2& 48.2& \textbf{23.8}& 64.9& 60.3&57.3\\
    \rowcolor{customgray!10}  \aname-2.7B-50\%   &1B& \textbf{86.8}&\textbf{75.1}& \textbf{59.5}& \textbf{64.6}& \textbf{35.1}& \textbf{48.7}& 22.4& \textbf{66.4}& \textbf{61.7}&\textbf{57.8}\\
     \rowcolor{customgray!20} GQA-2.7B-25\% &1B& 82.0& 72.8& 54.9& 58.4& 31.0& 42.9& \textbf{21.7}& 58.5& 49.6&52.4\\
     \rowcolor{customgray!20}  \aname-2.7B-25\% &1B& \textbf{85.6}& \textbf{74.1}& \textbf{57.6}& \textbf{61.5}& \textbf{32.4}& \textbf{45.9}& \textbf{21.7}& \textbf{63.1}& \textbf{56.9}&\textbf{55.4}\\
    \midrule
    \midrule
     S.-LLaMA-1.3B$^\dagger$ &2T& 87.0& 73.6& 58.2& 60.9& 29.5& 45.4& 21.8& 65.5& 61.3&55.9\\
    \rowcolor{customgray!10}  GQA-1.3B-50\%          &1B& 84.3& \textbf{72.3}& \textbf{55.8}& 57.5& 28.2& 41.8& 20.7& 62.9& 52.9&52.9\\
    \rowcolor{customgray!10}  \aname-1.3B-50\%   &1B& \textbf{84.5}& 72.0& 55.2& \textbf{58.1}& \textbf{28.7}& \textbf{42.6}& \textbf{21.5}& \textbf{63.7}& \textbf{55.4}&\textbf{53.6}\\
     \rowcolor{customgray!20} GQA-1.3B-25\% &1B& 76.6& 70.0& 52.9& 51.9& 23.5& 37.6& 21.0& \textbf{59.9}& 41.0&48.3\\
     \rowcolor{customgray!20} DHA-1.3B-25\% &1B& \textbf{82.8}& \textbf{71.1}& \textbf{54.0}& \textbf{55.4}& \textbf{25.8}& \textbf{40.5}& \textbf{21.5}& 57.6& \textbf{48.6}&\textbf{50.8}\\
     \bottomrule
    \end{tabular}
    }
    \label{tab:main_result}
    \vspace{-4mm}
\end{table}

%% file: Tables/ablation.tex
\begin{table}[t]
\centering
\small
\caption{\small Ablation Results of DHA \textit{w.o.} Linear Heads Fusion and Adaptvie Transformation. Experiments are conducted with LLaMA2-$7$B with $25\%$ heads budget and 0.5B \& 1B training budget on 0-shot Evaluation.}
\resizebox{\linewidth}{!}{
\begin{tabular}{cccccccccc}
\toprule
\textbf{Models} & \textbf{SciQ} & \textbf{PiQA} & \textbf{Wino.}  & \textbf{ARC-E.} & \textbf{ARC-C.} & \textbf{LogiQA} & \textbf{LAMB.} & \textbf{Average} & \textbf{Diff}  \\ \midrule
\aname-7B-25\% (0.5B)   & 88.6& 75.9& 61.3& 68.2& 36.1&23.8& 63.2& 59.6& $-$\\ \midrule
\rowcolor{customgray!10} \textit{w.o.} Linear Heads Fusion  & 83.4& 73.7& 57.3& 63.6& 29.4&  22.0& 51.9& 54.5& $-$5.1\\
\rowcolor{customgray!10} \textit{w.o.} Adaptvie Transformation  & 87.9& 74.1& 60.1& 69.4& 34.7&  19.5& 62.1& 58.3& $-$1.3\\
\midrule 
\midrule 
\aname-7B-25\% (1B)   & 90.0& 75.2& 63.8& 70.4& 37.5&21.1& 62.9& 60.1& $-$\\ \midrule
\rowcolor{customgray!10} \textit{w.o.} Linear Heads Fusion  & 87.5& 74.5& 60.7& 67.3& 32.8&  21.7& 58.3& 57.5& $-$2.6\\
\rowcolor{customgray!10} \textit{w.o.} Adaptvie Transformation  & 89.5& 74.6& 62.8& 69.1& 36.3&  21.6& 62.4& 59.5& $-$0.6\\
\midrule 
\midrule 
\aname-7B-25\% (5B)   & $91.7$ & $76.8$ & $64.4$ & $70.9$ & $42.8$ &$21.8$ & $68.4$ & 62.4 & $-$\\ 
\midrule 
 \rowcolor{customgray!20} GQA-7B-25\% (5B) & 91.5 &	76.6&	63.9& 70.5& 42.3&  22.1& 67.8& 62.1& $-$0.3\\
\bottomrule
\end{tabular}}
\label{tab:ablation}
\vspace{-2mm}
\end{table}

%% file: Figures/visual.tex
\begin{figure}[!t]
\vspace{-4mm}
    \centering
    \begin{minipage}{0.5\textwidth}
        \vspace{3mm}
        \centering
        \includegraphics[width=\textwidth]{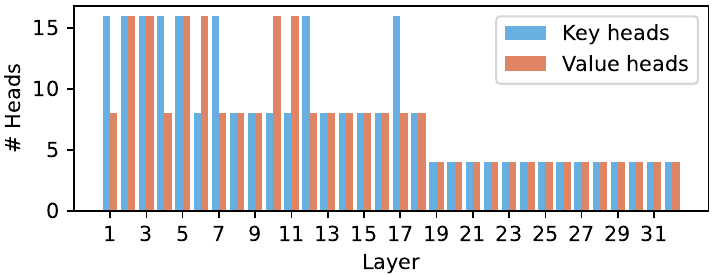}
        \caption{Allocation of key-value head budgets with 32 layers in DHA-7B-25\% after 240 step Search.}
        \label{fig:dha_budget}
    \end{minipage}\hfill
    \begin{minipage}{0.45\textwidth}
        \centering
        \includegraphics[width=\textwidth]{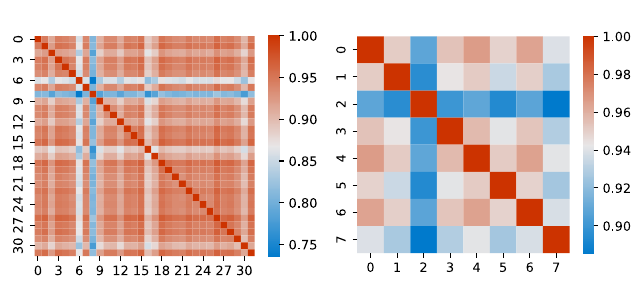}
        \caption{ Similarity of value heads in the 7th layer of LLaMA2-7B MHA \textbf{(Left)} or the transformed DHA-7B-25\% DHA \textbf{(Right)}.}
        \label{fig:prepost}
    \end{minipage}\hfill
  \vspace{-3mm}
\end{figure}

%% file: Tables/fusion_pt.tex
\begin{wraptable}[10]{ht}{.4\textwidth}
  \centering
    \small
    \vspace{-5mm}
    \caption{Data budget allocation to fusion and continued pre-training(CT) and 0-shot Task Average Accuracy (\%) in DHA-1.3B.}
    \resizebox{0.38\textwidth}{!}{%
    \begin{tabular}{cccc}
    \toprule
    \multicolumn{2}{c}{\textbf{Fusion}} & \multicolumn{2}{c}{\textbf{CT}} \\ \cmidrule(lr){1-2} \cmidrule(lr){3-4}
    Tokens & Avg.Acc & Tokens & Avg.Acc \\ \cmidrule(lr){1-2} \cmidrule(lr){3-4}
    0.05B                    & 33.74                       & 4.95B              & 59.08            \\
    \textbf{0.10B}                    & \textbf{38.32}                       & \textbf{4.90B}& \textbf{59.53}           \\
    0.15B                    & 48.26                       & 4.85B& 59.46            \\
    0.20B                    & 52.54                       & 4.80B& 59.16            \\ \bottomrule
    \end{tabular}
    }
    \label{tab:pruning_finetuning}
    \vspace{10mm}
\end{wraptable} 




%% file: related_work.tex

\vspace{-1mm}\paragraph{Advanced Multi-Head Attention.} Some efforts have been converting the traditional Multi-Head Attention (MHA)~\cite{vaswani2017attention} to Multi-Query Attention (MQA)~\cite{shazeer2019fast}, Group-Query Attention (GQA)~\cite{ainslie2023gqa} or GQKVA~\cite{javadi2023gqkva}. These methods achieve a balance between performance and efficiency by reducing the number of head parameters through parameter reuse across grouped heads. \aname is inspired by these methods and has a much higher optimization rate and much less training overhead.

\vspace{-1mm}\paragraph{Efficient Pre-training Approaches.} In recent years, the ability of incremental training to accelerate large-scale model training by studying how to obtain the optimal initialization point for training has thus attracted much attention ~\cite{DBLP:conf/cvpr/XieXP17,DBLP:conf/nips/WuW019}. 
Net2Net~\cite{chen2015net2net} uses function-holding transformations to expand the width by duplicating neurons, and uses a unitary layer implementation to expand the depth. LiGO~\cite{wangLearningGrowPretrained2023} proposes a learnable expansion method that can be used at the initial initialization point of a transformer. \aname is inspired by these methods, but we investigate how to learn to map the parameter matrix from large to small without losing the ability of the larger model itself. For additional related work, please refer to Appendix~\ref{app:related_work}.\vspace{-1mm}

%% file: appendix.tex
\section{Extended Related Works, Discussions, and Limitations}
\label{app:related_work}

\subsection{Extended Related Works}

\paragraph{Efficient Transformers.} EfficientTransformers\cite{tay2020efficient} have been extensively explored~\cite{sparse, kitaev2020reformer, bigbird, beltagy2020longformer, transformer_xl, erniedoc, rmt, chevalier2023adapting} to address the self-attention operation which scales quadratically with the sequence length. For instance, Sparse Transformer~\cite{sparse} uses a dilated sliding window the reduces the attention complexity. Longformer~\cite{beltagy2020longformer} and Bigbird~\cite{bigbird} reduced the complexity of self-attention by combining random, window and global attention. Recurrence Transformers~\cite{transformer_xl} maintain a memory bank of past KV cache to process the long text in segments. However, the above methods either result in a loss of model performance or require retraining the model, which is unaffordable for the high computational resources of LLMs. \aname requires very little computation to transform checkpoints into an efficient architecture that balances performance and computational resources.
\paragraph{KV Cache Compression.} KV Cache Compression methods emerged for reducing the prominent inference bottleneck caused by KV cache, particularly for long content input. A series of methods~\cite{h2o, ge2023model, liu2023scissorhands, msrnn, chen-etal-2024-nacl} explored the sparsity among Transformer's attention block, then evicted unnecessary tokens from KV Cache for efficient inference. 
However, these methods discard information from the context and use algorithms for inference that are inconsistent with the training phase, which can cause model performance degradation. \aname does not need to discard information from the context and is able to maintain consistent performance for training and inference.
Pruning~\cite{wangStructuredPruningLarge2020a, xiaShearedLLaMAAccelerating2023}, quantization~\cite{dettmers2022llmint8,frantar2023gptq,gray1998quantization} and distillation~\cite{hinton2015distilling,Gou_2021,cai2022pile} can reduce the number of model key and value headers, parameter dimensions, and activation to reduce memory bandwidth overhead during model inference. Deja Vu~\cite{pmlr-v202-liu23am} and CHAI~\cite{agarwal2024chai} prune pruning redundant heads through clustering methods for efficient inference. In the LLM era, this leads to a significant reduction in neuron redundancy as models move from task-specific to generalized~\cite{frantarSparseGPTMassiveLanguage2023}. The application of these methods to LLMs is computationally expensive and leads to performance degradation at larger pruning magnitudes. 

\textbf{Model Compression.}
Our approach is dedicated to obtaining a high-performance lightweight language model, which is the same goal as the task of model compression. Quantization~\cite{gray1998quantization} reduces the numerical accuracy of model weights and activations, and speeds up training and inference, but results in a loss of model accuracy and the inability to freely build target-specific models. CRash~\cite{zhangCRaShClusteringRemoving2023} and LayerDrop~\cite{zhang2020accelerating,sajjad2023effect} methods discard ineffective layers during training, which do not allow for target-specific structuring and come with a large performance loss. Pruning~\cite{wangStructuredPruningLarge2020a} minimizes the impact on performance by cutting out redundant neurons that over-parameterize the model. In the LLM era, this leads to a significant reduction in neuron redundancy as models move from task-specific to generalized~\cite{frantarSparseGPTMassiveLanguage2023}. Pruning LLM leads to performance degradation at larger pruning magnitudes. LLMsheairng~\cite{xiaShearedLLaMAAccelerating2023} uses the results of pruning as initialization for continuous pre-training of the model to recover performance, but this approach requires more data and computational overhead. We avoid the information loss caused, by learning the parameter fusion matrix of the model to reach a specific structure, thus obtaining better initialization points and reducing the overhead of continuous pre-training.

\subsection{Broader Impact and Limitations} \label{app:limitation}

\subsubsection{Broader Impact}
In this paper, we observe the MHA head mechanism and report the phenomenon of modular clustering of heads in MHA. This paper innovatively proposes linearly fusible parameters within the model, and designs linear fusion operators and related experiments to verify the low-loss fusible nature of the parameters. This helps to advance parameter fusion theory and LLM interpretability studies, which provide a foundation and inspiration for future algorithmic advancements, encouraging further optimization and innovation in LLMs. Our work on Decoupled-Head Attention (DHA) represents an advancement in optimizing the efficiency of Large Language Models (LLMs). By addressing the substantial computational and memory costs associated with the widely used Multi-Head Attention (MHA), DHA enhances the applicability of LLMs in various domains. The introduction of DHA not only achieves a remarkable balance between performance and efficiency but also significantly reduces the need for extensive pre-training, making the deployment of LLMs more feasible and cost-effective. This efficiency allows for the broader accessibility of advanced LLMs, democratizing technology and fostering innovation across industries. Furthermore, by requiring only 0.25\% of the original model's pre-training budgets to achieve near-original performance while saving 75\% of KV cache, DHA contributes to significant energy savings, aligning with sustainable and environmentally friendly AI practices. The enhanced performance and reduced training costs accelerate the development of AI applications, enhancing productivity in fields such as natural language processing, healthcare, and finance. 

\subsubsection{Limitation}
There are two limitations to our current approach. Firstly, we have only utilized linear methods for parameter fusion in our model. Future research should explore nonlinear methods, as they may offer a better way to link different parameters and achieve optimal results. Secondly, due to computational resource constraints, we have only experimented with models of 7 billion, 3 billion, and 1.3 billion parameters. However, our method is scalable and can be extended to models of any size in future work.

\subsubsection{Ethical Consideration}
In our study, we utilize publicly available data and techniques to address privacy concerns. Our approach focuses on improving model parameter efficiency and reducing model size to develop robust, compact, and accessible models, thus promoting the open dissemination and democratization of NLP technologies. By implementing pre-training strategies, we aim to mitigate biases through comprehensive training on large datasets, contributing to ethical AI development that prioritizes transparency, efficiency, and bias reduction. Our work is dedicated to advancing accessible and efficient NLP technologies, fostering a more inclusive and automated future for AI.

\section{More Implementation Details}
\subsection{Head Similarity and MHA Redundancy}\label{app:cka}
Centered Kernel Alignment (CKA) is a statistical measure used to quantify the similarity between two sets of data representations. Unlike traditional correlation measures, CKA is designed to be invariant to orthogonal transformations and scaling of the data.

To calculate the similarity between two sets of representations using CKA, we employ a kernel function to map the original data into a higher-dimensional space, where the alignment of their central tendencies can be more easily measured. The CKA value ranges from 0 to 1, where 0 indicates no similarity and 1 indicates identical representations.

The mathematical formulation of CKA, when using a linear kernel, is given by the following equation:

\[ \text{CKA}(X, Y) = \frac{\|X^TY\|_F^2}{\sqrt{\|X^TX\|_F^2 \cdot \|Y^TY\|_F^2}} \] 

Here, \(X\) and \(Y\) are matrices whose columns are the vectors of the representations to be compared, \(\|\cdot\|_F\) denotes the Frobenius norm, and \(X^T\) and \(Y^T\) are the transposes of \(X\) and \(Y\), respectively. To mathematically define the redundancy of each layer based on the average similarity between heads, we follow these steps:

\begin{enumerate}
    \item Compute the similarity between heads: For each pair of heads within a given layer, calculate the similarity using the CKA formula.
    \item Compute the average similarity: Average the similarity scores of all pairs of heads to define the redundancy of the layer.
\end{enumerate}

\subsubsection{Compute Similarity Between Heads}

Consider a layer with \( H \) heads, where the parameters of each head are represented by the matrices \( \textbf{W}_i \) (e.g., \( \textbf{W}_{q1}, \textbf{W}_{q2}, \ldots, \textbf{W}_{qH} \) for query weights). For each pair of heads \( i \) and \( j \), compute the CKA similarity using the following formula:

\[ \text{CKA}(\textbf{W}_i, \textbf{W}_j) = \frac{\|\textbf{W}_i^T \textbf{W}_j\|_F^2}{\sqrt{\|\textbf{W}_i^T \textbf{W}_i\|_F^2 \cdot \|\textbf{W}_j^T \textbf{W}_j\|_F^2}} \]

\subsubsection{Compute Redundancy}

Calculate the similarity for all pairs of heads and then compute the average similarity:

\[ \text{Redundancy} = \frac{2}{H(H-1)} \sum_{i=1}^{H-1} \sum_{j=i+1}^{H} \text{CKA}(\textbf{W}_i, \textbf{W}_j) \]

The coefficient \( \frac{2}{H(H-1)} \) ensures that the average similarity is computed over all pairs of heads. This redundancy measure reflects the degree of similarity between the parameters of different heads within each layer. A higher redundancy indicates that the parameters of different heads are more similar, implying a higher level of redundancy.


\subsection{Implementation in LLaMA2 Model} \label{app:implement}
Our method can theoretically transform MHA architecture  in any transformer model to efficient DHA architecture. Using LLaMA models as case studies, we implemented DHA transformation with various compression rates on all MHA layers. Only a very small number of additional parameters need to be introduced,  DHA significantly accelerates training and improves performance.

DHA adaptively gives search heads and heads connectivity relationship with redundancy in each MHA layer. Thus DHA assigns different group sizes at different layers and aggregates similar heads into one group to speed up fusion and reduce knowledge loss due to noise in fusion. As Shown in Fig~\ref{fig:DHA}, the transformation process of MHA to DHA can be divided into three stages.  

In order to keep the performance of the DHA model at the fusion start consistent with the MHA model, we initialize the operators of the DHA model to the MHA model with the corresponding scaling factors of query-key, query-value set to 1, and the corresponding scaling factors within the rest of the groups set to 0. At the beginning of every fusion process (e.g. $2\times, 4\times, 8\times$), the algorithm first performs multiple STEPs constrained only by the $\mathcal{L}_{\text{lm}}$ constraints to propagation, computing the $\mathcal{L}_{\text{fusion}}$ but not optimizing the linear fusion operator based on it. Based on the $\mathcal{L}_{\text{fusion}}$ between head and head as a measure of the distance between head and head we perform head clustering with the goal of minimizing the average loss of heads within each group and maximizing the average loss of heads between groups and groups. Afterwards, we select multiple groups with the smallest loss based on the compression rate as the fusion target, and optimize their $\mathcal{L}_{\text{fusion}}$ for back propagation. This algorithm ensures that the most redundant components of the model are fused and compressed during each transformation, while components requiring more parameters retain their original properties.

Our approach is theoretically applicable to transforming parameters across various transformer model designs, focusing on preserving the knowledge within MHA parameters.

Using LLaMA models as a case study, we implement our DHA transformation on all MHA layer. The whole transformation process can be divided into two phases: the Fusion phase with a small training budget and the recovery phase with continuous pre-training. Before Fusion phase, we define the total number of compressed headers budget $C$ then $C$ is split into compression rates at different compression levels. During Fusion phase, we modified the forward propagation path of MHA in the form of DHA refer to Eq.~\ref{eq:dhainit} and optimize $\mathcal{L}_{\text{fusion}}$ refer to Eq.~\ref{eq:dhaloss}. At the beginning, the fusion operators of each layer will be initialized making the DHA and the original MHA functionally equivalent. As we antagonistically optimize the fusion operator and upadte Lagrangian multipliers $\lambda$, the $\mathcal{L}_{\text{fusion}}$ that marks this DHA fusion process decreases.

When $\mathcal{L}_{\text{fusion}}$ is less than 1e-3 we terminate the fusion algorithm and enter the post-processing phase. The fusion weights within each group are computed by averaging the weights corresponding to each query-key and query-value within the group.We construct new DHA heads' parameters from the original MHA heads based on the fusion operator. After that, the fused model parameters can recover the performance and complete the transformation with a small amount of restorative pre-training.

We implemented the DHA algorithm with different compression ratios on models of different sizes. Experiments show that the DHA algorithm is adapted to models of various sizes. Only a very small number of additional parameters need to be introduced, and DHA preserves parameter knowledge in the model and improves performance.

\subsubsection{Attention Module Initialization}

In the module initialization process, the input key and value tensors are first reshaped and grouped according to the number of key and value heads, respectively. Given the batch size (bsz), number of heads (num\_heads), key length (k\_len), and head dimension (head\_dim), the key tensor is reshaped into keys\_grouped of shape [bsz, num\_key\_heads, num\_heads // num\_key\_heads, k\_len, head\_dim]. Similarly, the value tensor is reshaped into values\_grouped of shape [bsz, num\_value\_heads, num\_heads // num\_value\_heads, k\_len, head\_dim]. These grouped tensors are then expanded by repeating them along the group size dimension, resulting in keys\_expanded and values\_expanded. Correspondingly, the weight tensors weights\_k and weights\_v are reshaped to match the expanded dimensions and are then multiplied element-wise with the expanded key and value tensors.

\begin{algorithm}[h]
\caption{Attention Module Initialization}
\begin{algorithmic}[1]
\Require $K$ \Comment{key tensor}
\Require $V$ \Comment{value tensor}
\Ensure $K'$ \Comment{weighted key tensor}
\Ensure $V'$ \Comment{weighted value tensor}

\State $b \gets \text{batch size}$
\State $H \gets \text{number of heads}$
\State $L_k \gets \text{key length}$
\State $D \gets \text{head dimension}$

\State $K_s \gets \text{self.num\_key\_heads}$
\State $V_s \gets \text{self.num\_value\_heads}$

\State $K_g \gets \text{self.key\_group\_size}$
\State $V_g \gets \text{self.value\_group\_size}$

\State $K_g' \gets \text{self.weights\_k}$
\State $V_g' \gets \text{self.weights\_v}$

\State $K_g \gets K.\text{view}(b, K_s, H / K_s, L_k, D)$
\State $V_g \gets V.\text{view}(b, V_s, H / V_s, L_k, D)$

\State $K_e \gets K_g.\text{repeat\_interleave}(K_g, \text{dim}=1)$
\State $V_e \gets V_g.\text{repeat\_interleave}(V_g, \text{dim}=1)$

\State $K_w \gets K_g'.\text{view}(1, H, K_g, 1, D)$
\State $V_w \gets V_g'.\text{view}(1, H, V_g, 1, D)$

\State $W_K \gets K_e \times K_w$
\State $W_V \gets V_e \times V_w$

\State $K' \gets W_K.\text{sum}(\text{dim}=2)$
\State $V' \gets W_V.\text{sum}(\text{dim}=2)$

\State \Return $K'$, $V'$
\end{algorithmic}
\end{algorithm}

\subsubsection{Attention Forward Pass}
During the forward pass, the reshaping and expansion of the key and value tensors are performed in a similar manner as in the initialization process but with parameters specific to the DHA fusion phase. The key tensor is reshaped into keys\_grouped of shape [bsz, dha\_warmup\_group\_num, num\_heads // dha\_warmup\_group\_num, k\_len, head\_dim] and the value tensor into values\_grouped of shape [bsz, dha\_warmup\_group\_num, num\_heads // dha\_warmup\_group\_num, k\_len, head\_dim]. These grouped tensors are then expanded by repeating them according to the dha\_warmup\_group\_size. The weights weights\_k and weights\_v are reshaped and expanded to align with the dimensions of the expanded key and value tensors. Element-wise multiplication is performed between the expanded tensors and their corresponding weights, and the resulting weighted tensors are summed along the appropriate dimension.
\begin{algorithm}[h]
\caption{Attention Forward Pass}
\begin{algorithmic}[1]
\Require $K$ \Comment{key tensor}
\Require $V$ \Comment{value tensor}
\Ensure $K'$ \Comment{weighted key tensor}
\Ensure $V'$ \Comment{weighted value tensor}

\State $b \gets \text{batch size}$
\State $H \gets \text{number of heads}$
\State $L_k \gets \text{key length}$
\State $D \gets \text{head dimension}$

\State $G_q \gets \text{self.dha\_warmup\_group\_num}$
\State $G_s \gets \text{self.dha\_warmup\_group\_size}$

\State $K_g' \gets \text{self.weights\_k}$
\State $V_g' \gets \text{self.weights\_v}$

\State $K_g \gets K.\text{view}(b, G_q, H / G_q, L_k, D)$
\State $V_g \gets V.\text{view}(b, G_q, H / G_q, L_k, D)$

\State $K_e \gets K_g.\text{repeat\_interleave}(G_s, \text{dim}=1)$
\State $V_e \gets V_g.\text{repeat\_interleave}(G_s, \text{dim}=1)$

\State $K_w \gets K_g'.\text{view}(1, H, G_s, 1, D)$
\State $V_w \gets V_g'.\text{view}(1, H, G_s, 1, D)$

\State $W_K \gets K_e \times K_w$
\State $W_V \gets V_e \times V_w$

\State $K' \gets W_K.\text{sum}(\text{dim}=2)$
\State $V' \gets W_V.\text{sum}(\text{dim}=2)$

\State \Return $K'$, $V'$
\end{algorithmic}
\end{algorithm}

\subsubsection{DHA Loss Calculation}

The calculation of the loss function in this model involves the adaptive DHA loss. This loss is computed based on the global step, warmup steps, and a base value. The DHA margin is calculated as the product of an exponential decay term and a linear decay term, ensuring it is non-negative. The adaptive DHA loss is derived by comparing the mean squared error (MSE) with the DHA margin and summing the positive differences.

Formally, the DHA margin \( M_{\text{dha}} \) is calculated as:

\[
M_{\text{dha}} = \max\left(0, \left( \text{base}^{\text{global\_step}} \right) \times \left( 1.0 - \frac{\text{global\_step}}{\text{dha\_warmup\_step}} \right) \right)
\]

MSE Loss are defined in Eq.~\ref{eq:dhaloss}. The adaptive DHA loss \( L_{\text{dha}} \) is then:

\[
L_{\text{dha}} = \sum \max(\text{mse} - M_{\text{dha}}, 0.0)
\]

The overall loss \( L \) is the adaptive DHA loss:

\[
L = L_{\text{dha}} + L_{\text{lm}}
\]

This combined loss function effectively utilizes the adaptive component to optimize the attention mechanism in the model. The calculation process ensures that the model adapts dynamically during training, reducing the loss progressively as the training steps increase.

\begin{algorithm}[H]
\caption{Adaptive DHA Loss Calculation}
\begin{algorithmic}[1]
\Require $B$ \Comment{blocks}
\Require $mse$ \Comment{Mean Squared Error tensor}
\Ensure $L$ \Comment{loss\_dha\_diversity}

\State $\lambda \gets \text{self.lambda\_mse}$
\State $s \gets \text{self.mse\_scale}$
\State $L \gets 0.0$
\State $global\_step \gets 1000$
\State $dha\_warmup\_step \gets 200$
\State $base \gets 0.999$

\For {each $b \in B$}
    \State $L_l \gets 0.0$ \Comment{loss\_dha\_diversity\_layer}
    \State $A \gets b.attn$ \Comment{attn\_layer}
    \State $W_k \gets A.weights\_k$
    \State $W_v \gets A.weights\_v$
    \State $G_s \gets A.dha\_warmup\_group\_size$
    \State $G_n \gets A.dha\_warmup\_group\_num$
    \State $H_{kv} \gets A.num\_key\_value\_heads$
    \State $H \gets A.num\_heads$
    \State $D \gets A.head\_dim$
    
    \State $W_k \gets \text{reshape}(W_k, [H_{kv}, -1, G_s, D])$
    \State $W_v \gets \text{reshape}(W_v, [H_{kv}, -1, G_s, D])$
    
    \For {each $r \in W_k$}
        \For {each $o \in W_k$ after $r$}
            \State $L_l \gets L_l + \text{MSE\_Loss}(r, o)$
        \EndFor
    \EndFor
    
    \For {each $r \in W_v$}
        \For {each $o \in W_v$ after $r$}
            \State $L_l \gets L_l + \text{MSE\_Loss}(r, o)$
        \EndFor
    \EndFor
    
    \State $N \gets H \times G_s \times D \times 2 \times \frac{(G_s - 1)}{2}$
    \State $L_l \gets \frac{s \times L_l}{N}$
    \State $L \gets L + L_l$
\EndFor

\State $L \gets \frac{L}{\text{len}(B)}$
\State $L \gets L \times \lambda$

\State $exponent \gets base^{global\_step}$
\State $linear\_decay \gets 1.0 - \frac{global\_step}{dha\_warmup\_step}$
\State $margin \gets \max(0, exponent \times linear\_decay)$
\State $adaptive\_loss \gets \sum \max(mse - margin, 0.0)$
\State $L \gets L + adaptive\_loss$

\State \Return $L$
\end{algorithmic}
\end{algorithm}

\subsection{Head Grouping Based on Fusion Loss} \label{app:cluster}

This algorithm uses simulated annealing to optimize group scores based on a given score matrix. It begins by defining the number of groups and distributing the points among them randomly. The initial score for these groups is calculated using the `calculate\_score` function, which sums the scores from the matrix for each group, considering each connection twice and dividing by two.

The algorithm starts with a high temperature (T=100) and gradually cools down (T\_min=0.001) using a cooling rate (alpha=0.9). During each iteration, two random points from different groups are swapped, creating a new grouping. The score for this new grouping is calculated, and the difference in score (delta) is evaluated.

If the new score is higher, or if a randomly generated number is less than the exponential of delta divided by the temperature, the new grouping is accepted. This allows the algorithm to escape local optima. The temperature is then reduced according to the cooling rate. This process continues until the temperature reaches the minimum threshold. The algorithm returns the final group configuration and its corresponding score, which represents an optimized grouping based on the initial score matrix.

In practice, we use the MSE computed by the head and the head as scores, and compute the matrix of scores between the head and the head for head clustering after forward.

\begin{algorithm}[H]
\caption{Head Grouping Optimization on Fusion Loss}
\begin{algorithmic}[1]
\Require $M$ \Comment{score matrix}
\Require $G \gets 8$ \Comment{number of groups}
\Ensure $best\_groups, best\_score$ \Comment{final groups and their score}

\Function{calculate\_score}{$M, groups$}
    \State $score \gets 0$
    \For {each $group \in groups$}
        \For {each $i \in group$}
            \For {each $j \in group$}
                \State $score \gets score + M[i][j]$
            \EndFor
        \EndFor
    \EndFor
    \State \Return $score / 2$ \Comment{each connection counted twice}
\EndFunction

\Function{simulated\_annealing}{$M, G$}
    \State $P \gets \text{length}(M)$ \Comment{number of points}
    \State $N \gets P / G$ \Comment{number of points per group}
    \State $points \gets \text{array}(range(P))$
    \State $\text{shuffle}(points)$
    \State $groups \gets points.reshape(G, N)$
    \State $current\_score \gets \Call{calculate\_score}{M, groups}$
    \State $T \gets 100.0$ \Comment{initial temperature}
    \State $T_{min} \gets 0.001$ \Comment{minimum temperature}
    \State $\alpha \gets 0.9$ \Comment{cooling rate}

    \While {$T > T_{min}$}
        \State $i, j \gets \text{random integers in } [0, G)$
        \If {$i \neq j$}
            \State $a, b \gets \text{random integers in } [0, N)$
            \State $new\_groups \gets groups.copy()$
            \State $temp \gets new\_groups[i][a]$
            \State $new\_groups[i][a] \gets new\_groups[j][b]$
            \State $new\_groups[j][b] \gets temp$

            \State $new\_score \gets \Call{calculate\_score}{M, new\_groups}$
            \State $\Delta \gets new\_score - current\_score$

            \If {$\Delta > 0 \textbf{ or } \exp(\Delta / T) > \text{random}()$}
                \State $groups \gets new\_groups$
                \State $current\_score \gets new\_score$
            \EndIf
        \EndIf

        \State $T \gets T \times \alpha$ \Comment{cooling down}
    \EndWhile

    \State \Return $groups, current\_score$
\EndFunction

\State $best\_groups, best\_score \gets \Call{simulated\_annealing}{M, G}$
\end{algorithmic}
\end{algorithm}

\subsection{Layer Allocation Based on Fusion Loss} \label{app:layerallo}

This algorithm efficiently allocates resources to different layers based on their respective losses to optimize system performance. Initially, it assigns a minimum allocation to each layer. Then, it calculates weights for each layer based on their losses, prioritizing layers with higher losses. The algorithm determines the number of times 16 can be allocated based on the remaining allocation. It allocates 16s to layers with the highest weights until reaching a predetermined limit. Next, it redistributes the remaining allocation to layers with the highest loss-to-allocation ratios, assigning resources in multiples of 8 or 4. This process ensures that layers with higher losses receive more resources, optimizing the overall system performance. Finally, the algorithm returns the final allocation for each layer, resulting in an efficient distribution of resources across the system. The total search process for the LLaMA2 model requires 42 minutes.

\begin{algorithm}[H]
\caption{Layer Allocation Based on Losses}
\begin{algorithmic}[1]
\Require $L$ \Comment{losses for each layer}
\Require $A \gets [4, 8, 16]$ \Comment{possible allocations}
\Require $T \gets 256$ \Comment{total allocation}
\Ensure $alloc \gets [a_1, a_2, \ldots, a_n]$ \Comment{final allocations for each layer}

\State $n \gets \text{length}(L)$
\State $alloc \gets [4] \times n$ \Comment{initial allocation}
\State $W \gets \frac{L}{\sum L}$ \Comment{weights proportional to losses}

\State $R \gets T - \sum alloc$ \Comment{remaining allocation}
\State $k \gets 1$ \Comment{initial number of 16's to allocate}
\State $M_{16} \gets R // 16$ \Comment{maximum number of 16's that can be allocated}
\State $k \gets \min(k, M_{16})$

\For {$i \gets 1$ to $k$}
    \State $idx \gets \text{argmax}(W)$
    \State $alloc[idx] \gets alloc[idx] + 12$
    \State $W[idx] \gets 0$ \Comment{prevent reallocation}
\EndFor

\State $R \gets T - \sum alloc$

\While {$R > 0$}
    \If {$R \geq 8$}
        \State $idx \gets \text{argmax}(\frac{L}{alloc})$
        \State $alloc[idx] \gets alloc[idx] + 4$
        \State $R \gets R - 4$
    \ElsIf {$R \geq 4$}
        \State $idx \gets \text{argmax}(\frac{L}{alloc})$
        \State $alloc[idx] \gets alloc[idx] + 4$
        \State $R \gets R - 4$
    \EndIf
\EndWhile

\State \Return $alloc$
\end{algorithmic}
\end{algorithm}
\newpage
\subsection{Training Details}
\label{app:throughput}
The hyperparameters used in our experiments are presented in Tab.~\ref{tab:training_details}. We employ fully sharded data parallel to efficiently train our models in parallel, and we utilize FlashAttention V1 \cite{dao2022flashattention} to accelerate the training process. A cosine learning rate scheduler is used, with the learning rate decaying to a minimum of 10\% of the peak value. Preliminary experiments were conducted to determine the optimal peak learning rate for learning the fusion variables and Lagrange multipliers. 

\begin{table}[h]
    
    \centering
    \caption{Training hyper-parameters}
    \begin{tabular}{lcc} \toprule
                                                                    & Fusion  & Contined Pre-training \\ \midrule
    Training budget                                                 & $0.2$B     & $5$B                   \\
    Learning rate of $\omega, \lambda$ & $0.05$        & -                     \\
    Learning Rate of $\theta$                          & $0.0001$ & $0.0001$              \\
    LR warmup ratio                                                 & $10\%$     & $3\%$                   \\
    Batch size        (tokens)                                              & $262$K     & $1$M                    \\
    Evaluation interval $m$       (steps)                                 & $40$       & $40$               \\ 
    Steps & $800$ & $5,000$ \\
    \# GPUs & $8$ & $8$ \\
    \bottomrule 
    \end{tabular}
    \label{tab:training_details}
\end{table}

\section{Extended Experiments}

\subsection{Instruction Tuning Evaluation.}\label{app:sft}

\paragraph{Instruction Tuning Evaluation.}
To assess our models' capabilities in downstream application after instruct tuning~\cite{ouyang2022instructgpt,alpaca}, we fine-tune both \aname and the baseline models on 10,000 instruction-response pairs drawn from the initial round of multi-turn chat histories in the ShareGPT dataset\footnote{\url{https://sharegpt.com}}. For evaluation, we select another 1,000 instructions from ShareGPT, generate responses using our fine-tuned models and other baseline models and employ GPT-4 as an evaluator to compare these responses \cite{dubois2023alpacafarm}. We report the win rate of our model relative to the baseline model.

\begin{figure}[h]
    \centering
    \includegraphics[width=\linewidth]{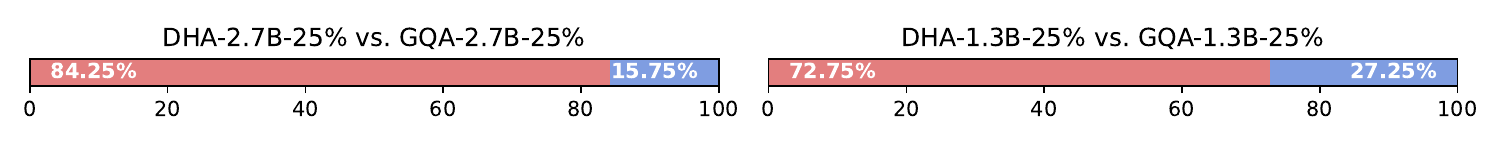}
    \caption{In model scale of 7B, 3B, and 1.3B, DHA significantly outperforms GQA and achieves comparable performance with MHA after instruction tuning .}
    
    \label{fig:win_rate}
\end{figure}

\paragraph{Instruction Tuning.} As shown in Fig.~\ref{fig:win_rate}, the tuned \aname model outperforms all GQA baselines of comparable scale . This demonstrates that the DHA model effectively retains the foundational capabilities of the MHA model and can be activated through instruction tuning to produce long, coherent, and informative responses.

\paragraph{Combination with KV Cache Compression Techniques.}

 In Sections~\ref{sec.bg} and~\ref{sec.method}, we demonstrated that \aname is a more efficient GQA architecture, so it has similarly good compatibility. We tested the compatibility of the DHA model with the KVCache eviction method NACL~\cite{chen-etal-2024-nacl}. NACL 25\% indicates retaining only 25\% of the KVCache. The experiment results are shown in the Tab.~\ref{tab.app.nacl}. DHA and GQA exhibit equally good compatibility with KV cache compression techniques.

\begin{table}[h!]
\centering
\caption{Comparison of log(PPL) between \aname and GQA with NACL.}
\begin{tabular}{lc}
\toprule
\textbf{Method} & \textbf{log(PPL)} \\ \midrule
GQA-7b-25\% & 2.89 \\ 
DHA-7b-25\% & 2.84 \\ \midrule
GQA-7b-25\% (NACL 25\%) & 3.01 \\ 
DHA-7b-25\% (NACL 25\%) & 2.93 \\ 
\bottomrule 
\end{tabular}
\label{tab.app.nacl}
\end{table}
\begin{table}[h!]
\centering
\caption{Comparison of Avg ACC and PPL between different methods at 7B-25\% (5B).}
\begin{tabular}{lcc}
\toprule
\textbf{Method} & \textbf{Avg ACC} & \textbf{PPL} \\ \midrule
DHA-7B-25\% (5B) & 62.4 & 7.29 \\ 
GQA-7B-25\% (5B) & 60.3 & 7.54 \\ 
GQA (CKA-Grouping)-7B-25\% (5B) & 60.4 & 7.51 \\ 
\bottomrule
\end{tabular}
\label{tab.app.gqaplus}
\end{table}

\paragraph{Compare with Advance GQA Initialization.} It's a common and effective approach to convert MHA to GQA using mean pooling instead of training from scratch. The author of GQA tested several methods for the initialization of GQA and found it works best using simple mean pooling from MHA. Indeed, training GQA from scratch will cost trillions tokens budget to match the performance of MHA which is inefficient and costly.Inspired by the similarity of head parameters, we improved the initialization method of GQA: instead of direct grouping, we first cluster similar heads using CKA and then perform mean-pooling initialization within each cluster. We compare this approach with the Vanilla GQA and DHA.

Tab.~\ref{tab.app.gqaplus} shows that GQA(CKA-Grouping)-7B-25\% (5B) achieved comparable performance to the original implementation in Vanilla GQA. We believe the reason for this is that the head grouping learned by DHA is based on the fusible nature between heads, which cannot be completely equated with CKA similarity. More importantly, DHA not only groups heads based on similarity but also learns the fusible parameters. This allows it to eliminate the influence of redundant parameters and retain more important information during the initialization process, which is not possible with mean initialization.

\section{Extend Analysis}

\paragraph{How Merging Weights Change.}
 Refer to Fig.~\ref{fig:loss_ratio}, where we show the weight variation diagram. In the fusion process of heads 0-3, head 0 initially constitutes 100\% as the starting head of the MHA. As the fusion process progresses, the parameters of the important heads increase, and the proportions of all heads become more balanced. This indicates that the algorithm attempts to retain information from different heads by balancing the parameter proportions of each head. This process results in a slight increase in loss, but not significantly.

\paragraph{\aname's Compatibility on GQA Model.} \aname is primarily designed for models based on the Transformer Decoder architecture and can be adapted to all models with this architecture. We chose LLaMA~\cite{240102415LLaMA} as the experimental baseline because it is a classic model using the decoder architecture in LLMs. Other open-source LLM models differ from LLaMA only in certain details (such as activation functions and training methods), which do not affect \aname's training. Successfully applying \aname to LLaMA indicates that it can be used in most decoder-only models. GQA~\cite{ainslie2023gqa} is an efficient variant of MHA, which optimizes the inference process through head grouping and sharing. Due to its simplicity and efficiency, GQA is widely used. DHA can be similarly constructed based on GQA, requiring only minor adjustments to the construction process. Here, we provide two feasible methods to convert GQA to \aname.

\begin{itemize}
    \item Easiest method in less than 1 minute. GQA can be losslessly converted into MHA by simply replicating the GQA' KV heads. Then, we can perform the \aname transformation on the MHA architecture.
    \item Minor modification by grouping KV. \aname only needs to group and fuse the Key and Value heads. When constructing \aname on GQA, we initially group the Key and Value, maintaining alignment with GQA functionality. During the training phase, the fused head parameters can replace the original GQA heads for sharing.
\end{itemize}

\paragraph{Inter-layer Grouping of Heads or Only Intra-layer Grouping?} Only intra-layer grouping and fusion is conducted in \aname. Fig.~\ref{fig:intro} meant to illustrate the decoupled-heads where the number of key and value heads can be different among layers. The \aname method employs parameter fusion within each layer for three reasons:

\begin{itemize}
    \item Higher redundancy of heads within layer for fusion. The heads within a layer exhibit high similarity and redundancy, which provides a good starting point for parameter fusion.
    \item More complex optimization for inter-layer fusion. The optimization process between layers is very complex and requires memory operations for cross-layer calls, which inherently increases the inference cost.
    \item Promising future work by introducing inter-layer fusion~\cite{chen-etal-2024-lemon}. This paper represents an early exploration of applying parameter fusion methods within model parameters. The inter-layer fusion approach is indeed a valuable direction for future exploration.
\end{itemize}

\paragraph{Accuracy Loss after Transformation.} The performance gap between the results shown in the paper and MHA is primarily due to the following two reasons:

\begin{itemize}
    \item The gap of pre-training data. The MHA model was not trained on the same data used for DHA. Since LLaMA's training data is not directly open-sourced, we used an experimental open-sourced pre-training data following Sheared-LLaMA (Xia et al., 2024). The improved pre-training data will close the gap between DHA and MHA.
    \item Parameter size difference. Compared to MHA, DHA compresses 50\% or 25\% of attention heads, requires only 0.05\% of pre-training data and achieves approximately 5\% loss. The number of parameters of MHA is much larger than that of DHA, so performance loss is inevitable during conversion. Compared with GQA, a strong baseline with the same number of parameters, DHA has shown higher training efficiency and performance advantages. Due to the high efficiency of DHA, DHA can use more heads than MHA with the same number of parameters, and has the opportunity to achieve better performance.
\end{itemize}

\newpage
\section{Extend Observation}

\subsection{Header parameter characteristics in MHA} \label{app:obsmha}
We show more of our head similarity observations in the LLaMA2-7b model MHA. Each subfigure represents the similarity between heads within the same layer for three different types of attention mechanisms: WQ (query), WK (key), and WV (value). The matrices are arranged in a 3x4 grid layout, with each row corresponding to a specific layer and each column corresponding to a type of attention mechanism. Note: Layer numbers start from 1.

\begin{figure}[htbp]
    \centering
    \begin{tabular}{cc}
        \begin{subfigure}[b]{0.47\textwidth}
            \centering
            \includegraphics[width=\textwidth]{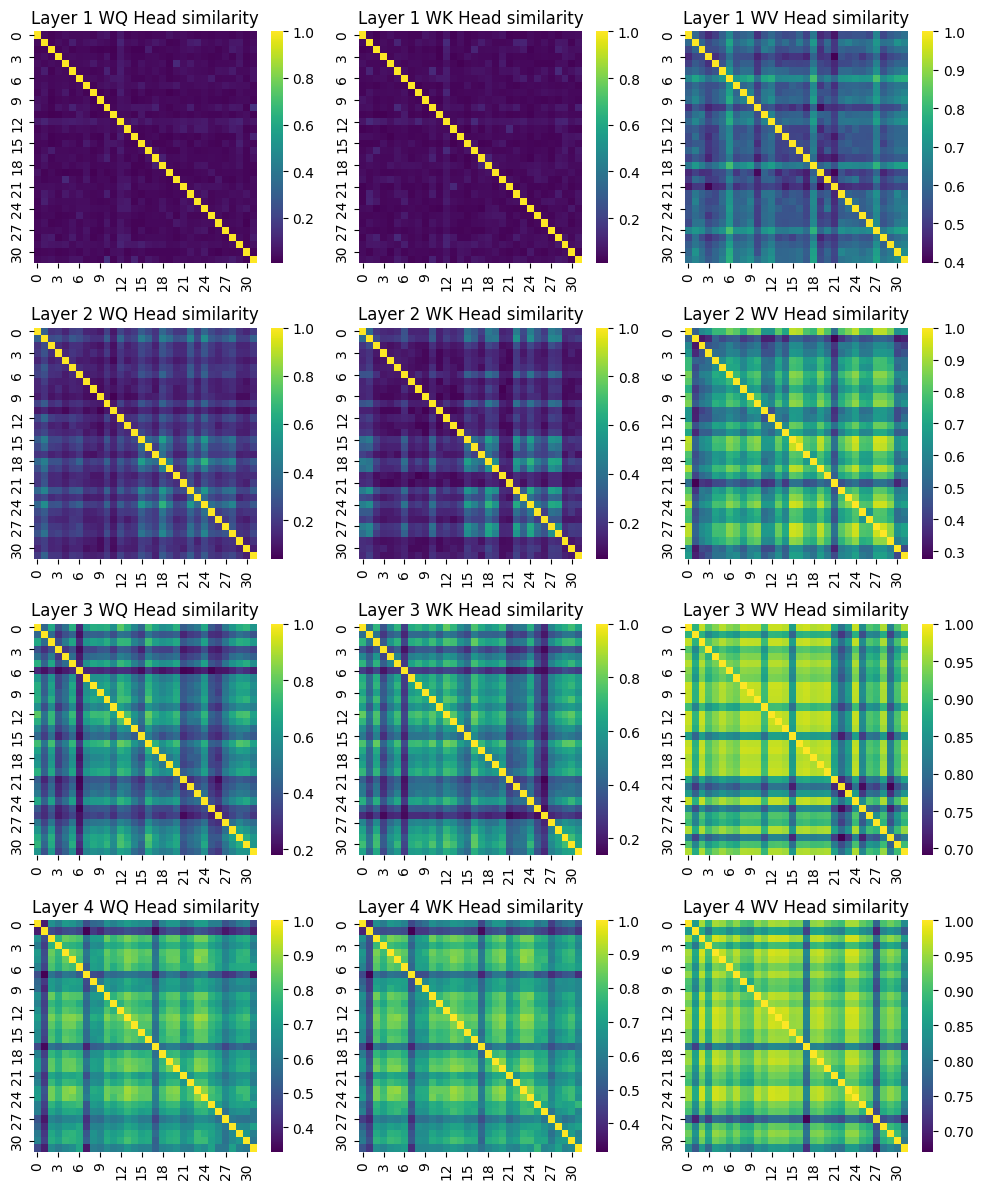}
        \end{subfigure} &
        \begin{subfigure}[b]{0.49\textwidth}
            \centering
            \includegraphics[width=\textwidth]{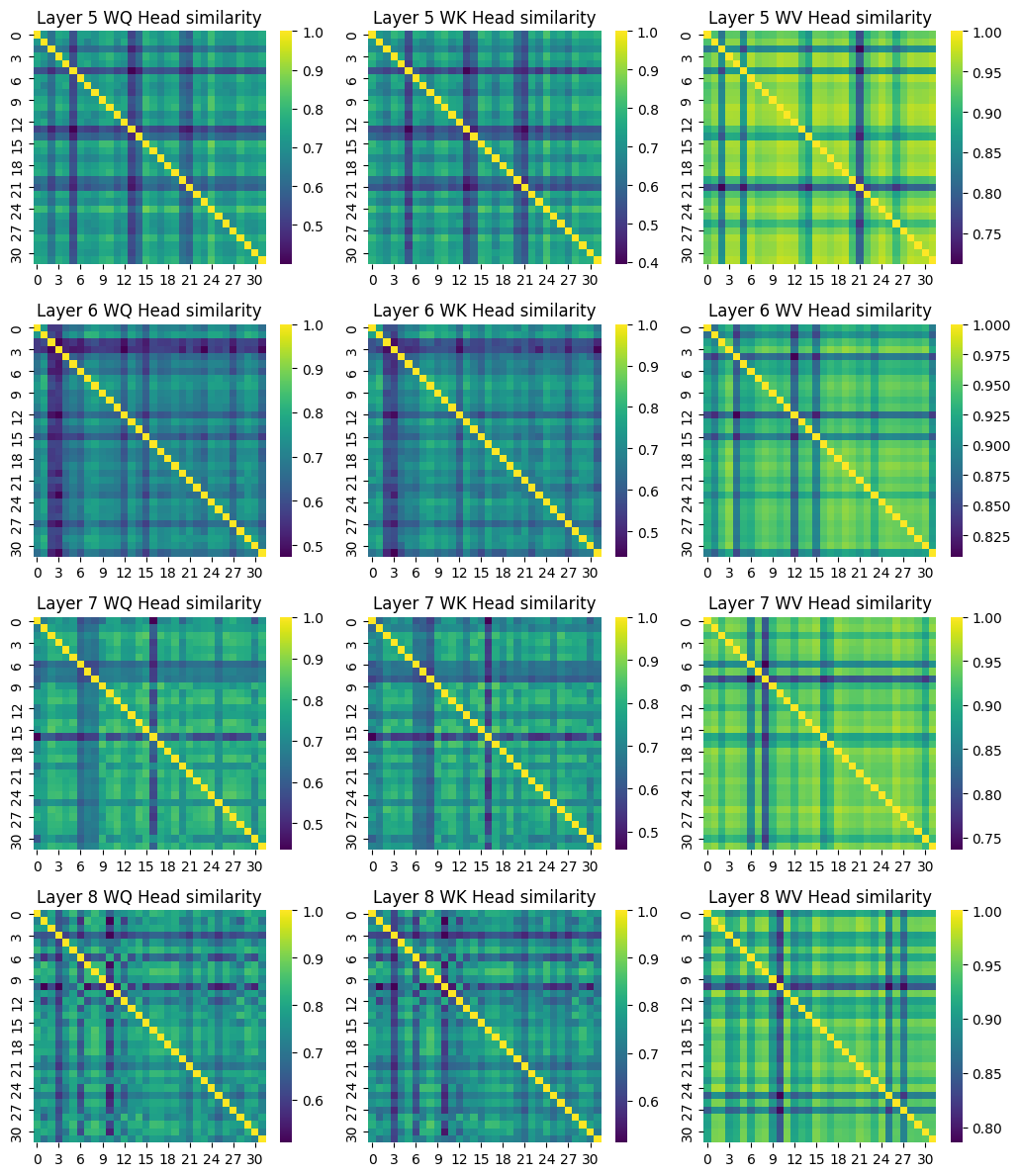}
        \end{subfigure} \\
        \begin{subfigure}[b]{0.49\textwidth}
            \centering
            \includegraphics[width=\textwidth]{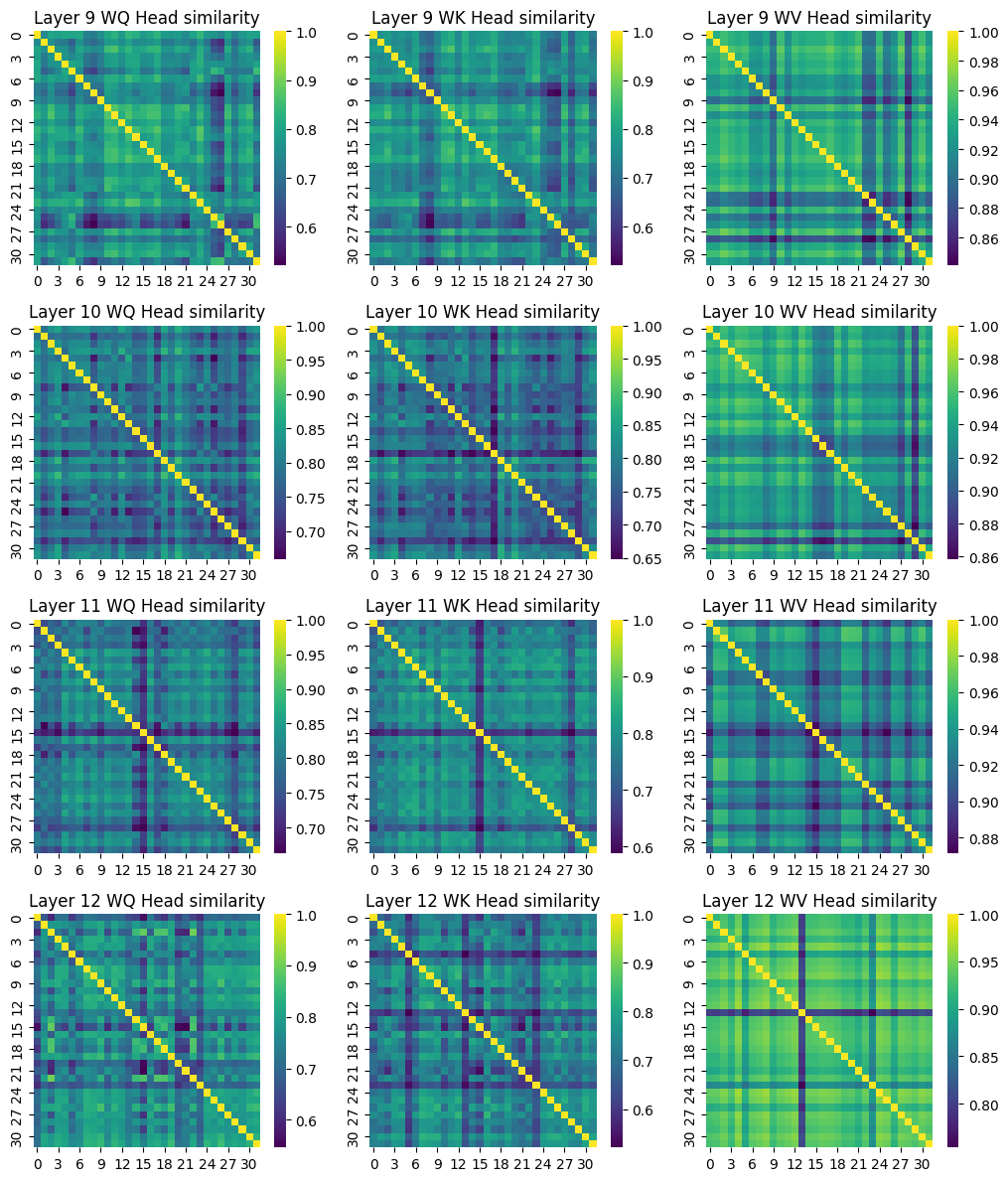}
        \end{subfigure} &
        \begin{subfigure}[b]{0.49\textwidth}
            \centering
            \includegraphics[width=\textwidth]{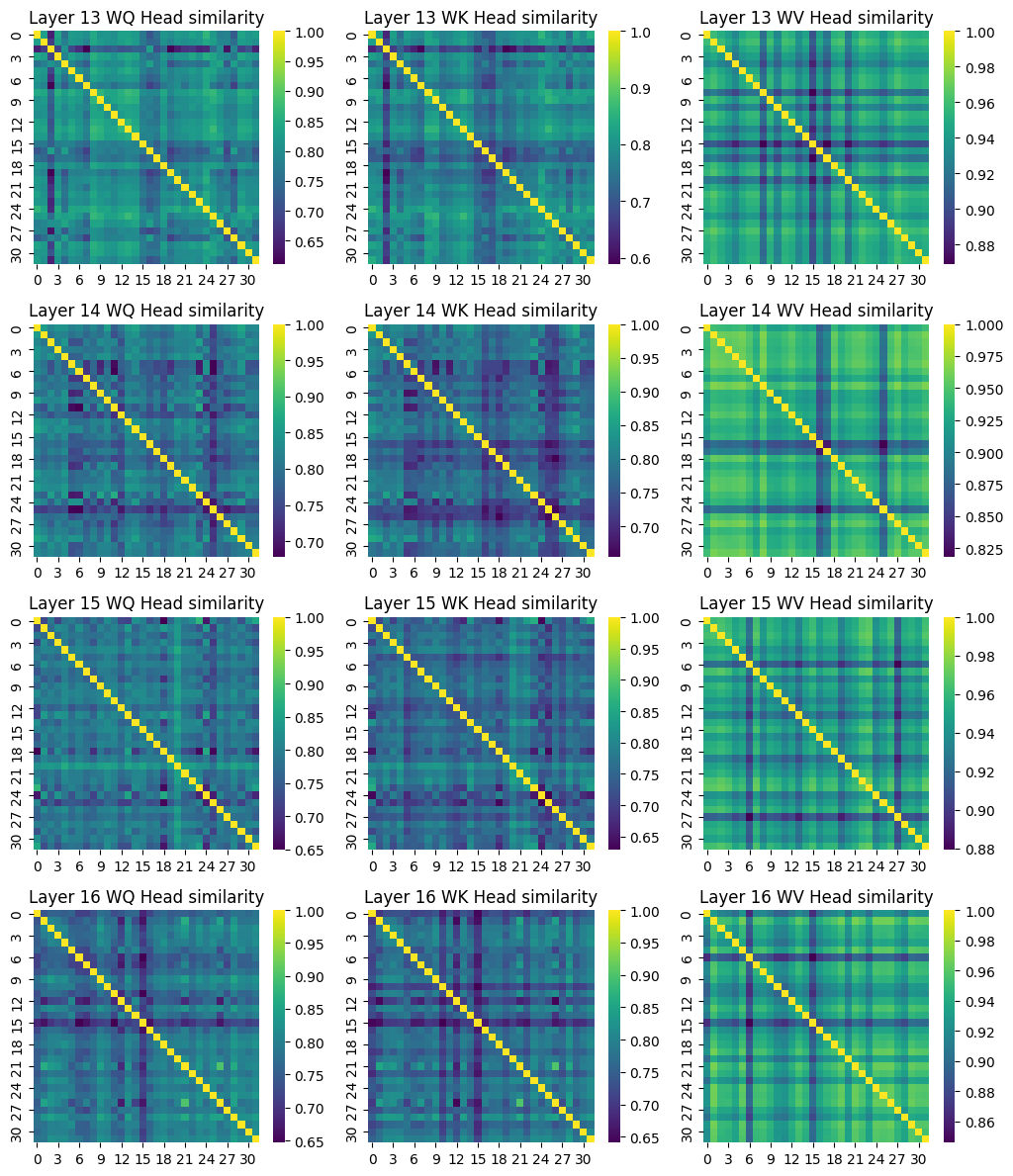}
        \end{subfigure}
    \end{tabular}
    \label{fig:2x2}
    \caption{Visualization of query, key, value head parameters similarity from layer 1 to layer 1b in LLaMA2-7B.}
\end{figure}

\newpage

\begin{figure}[htbp]
    \centering
    \begin{tabular}{cc}
        \begin{subfigure}[b]{0.49\textwidth}
            \centering
            \includegraphics[width=\textwidth]{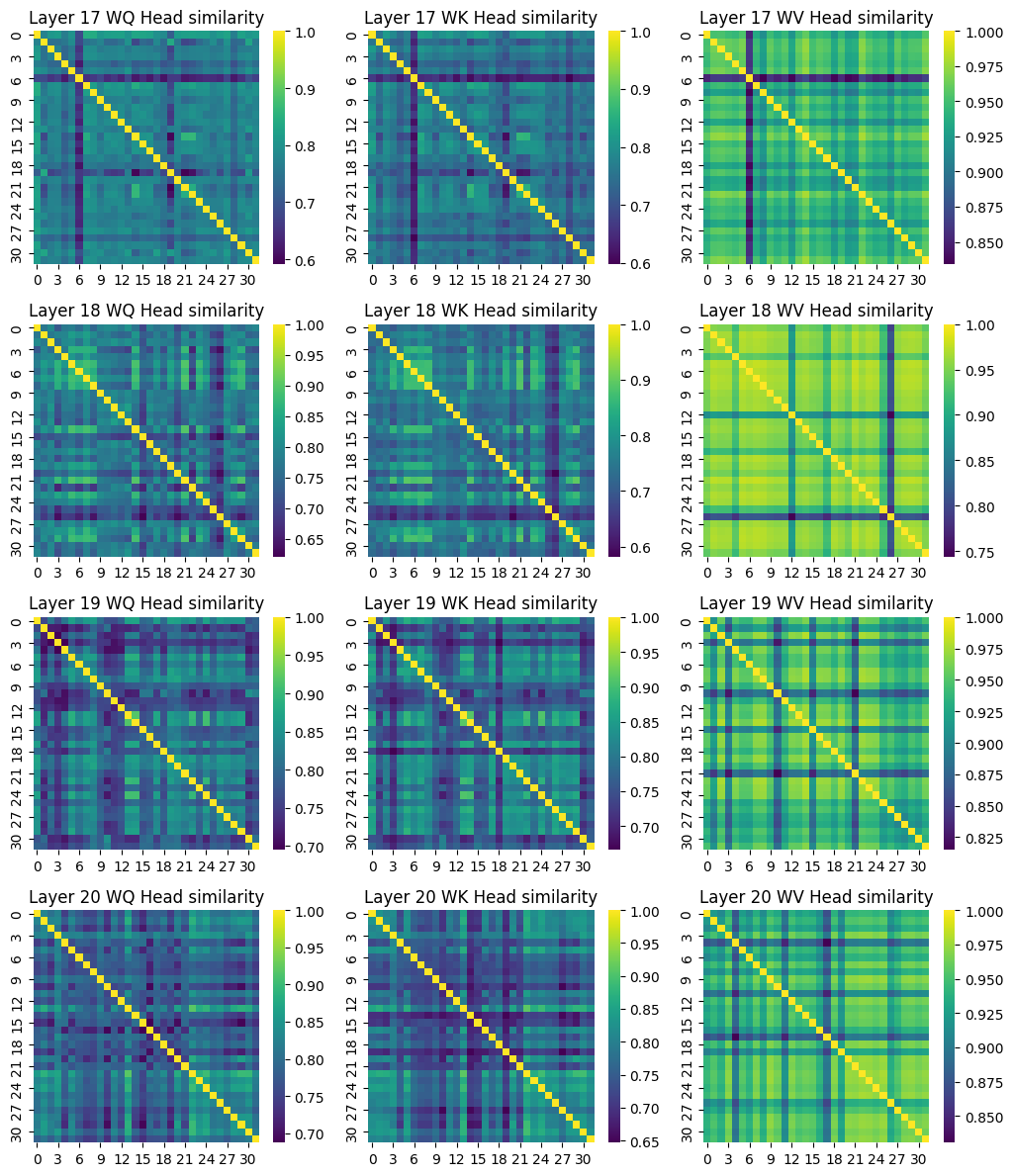}
        \end{subfigure} &
        \begin{subfigure}[b]{0.49\textwidth}
            \centering
            \includegraphics[width=\textwidth]{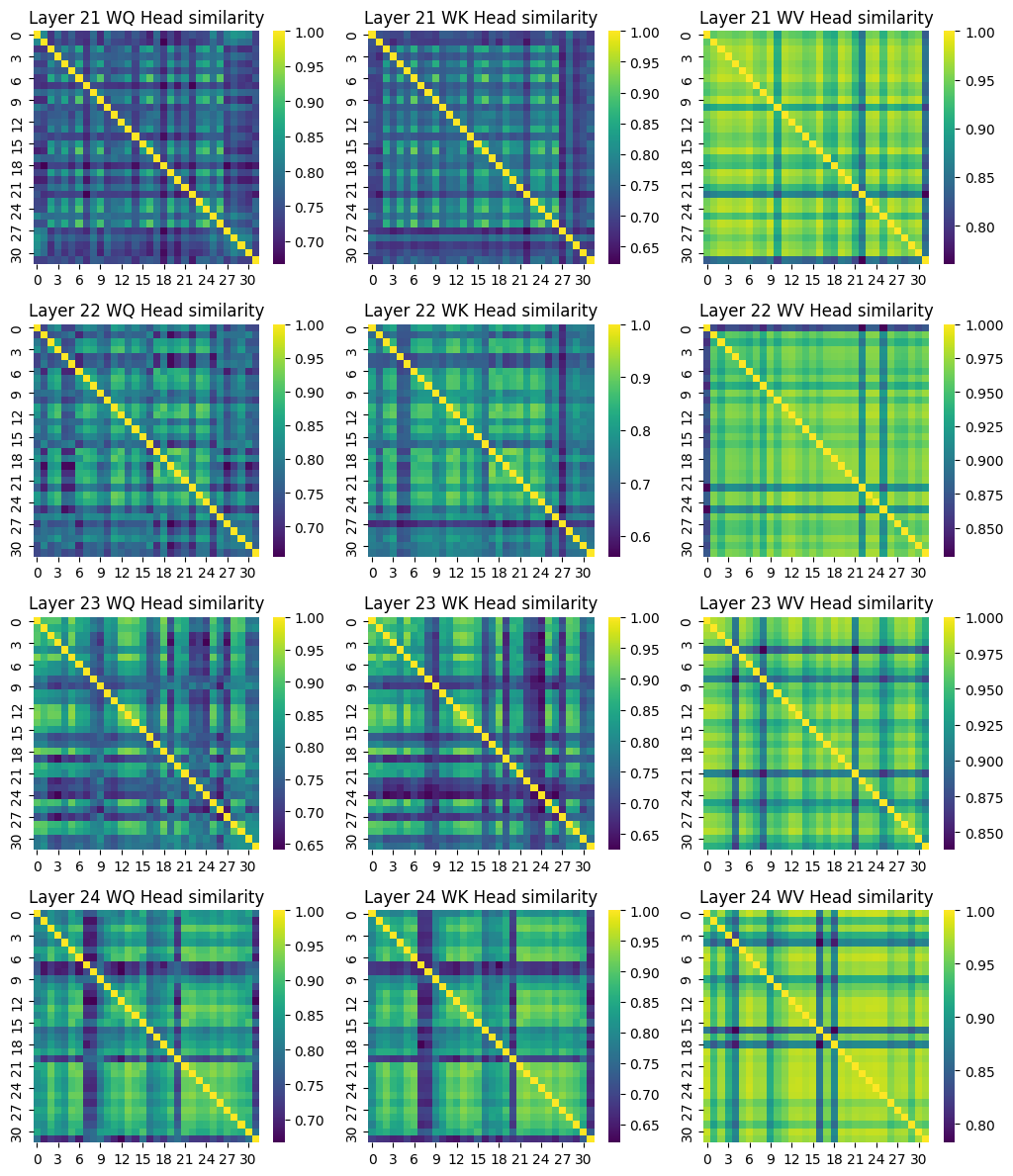}
        \end{subfigure} \\
        \begin{subfigure}[b]{0.49\textwidth}
            \centering
            \includegraphics[width=\textwidth]{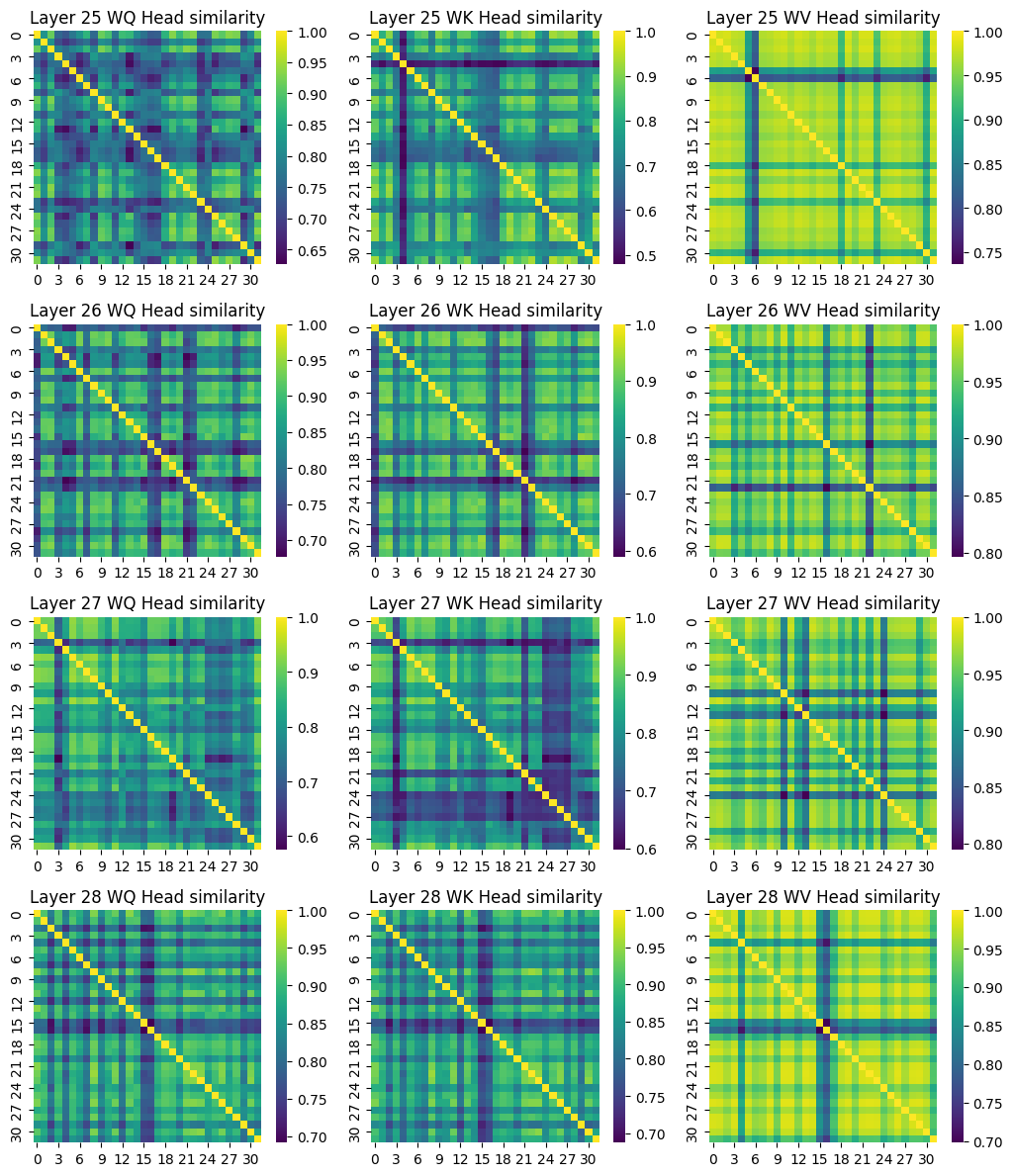}
        \end{subfigure} &
        \begin{subfigure}[b]{0.49\textwidth}
            \centering
            \includegraphics[width=\textwidth]{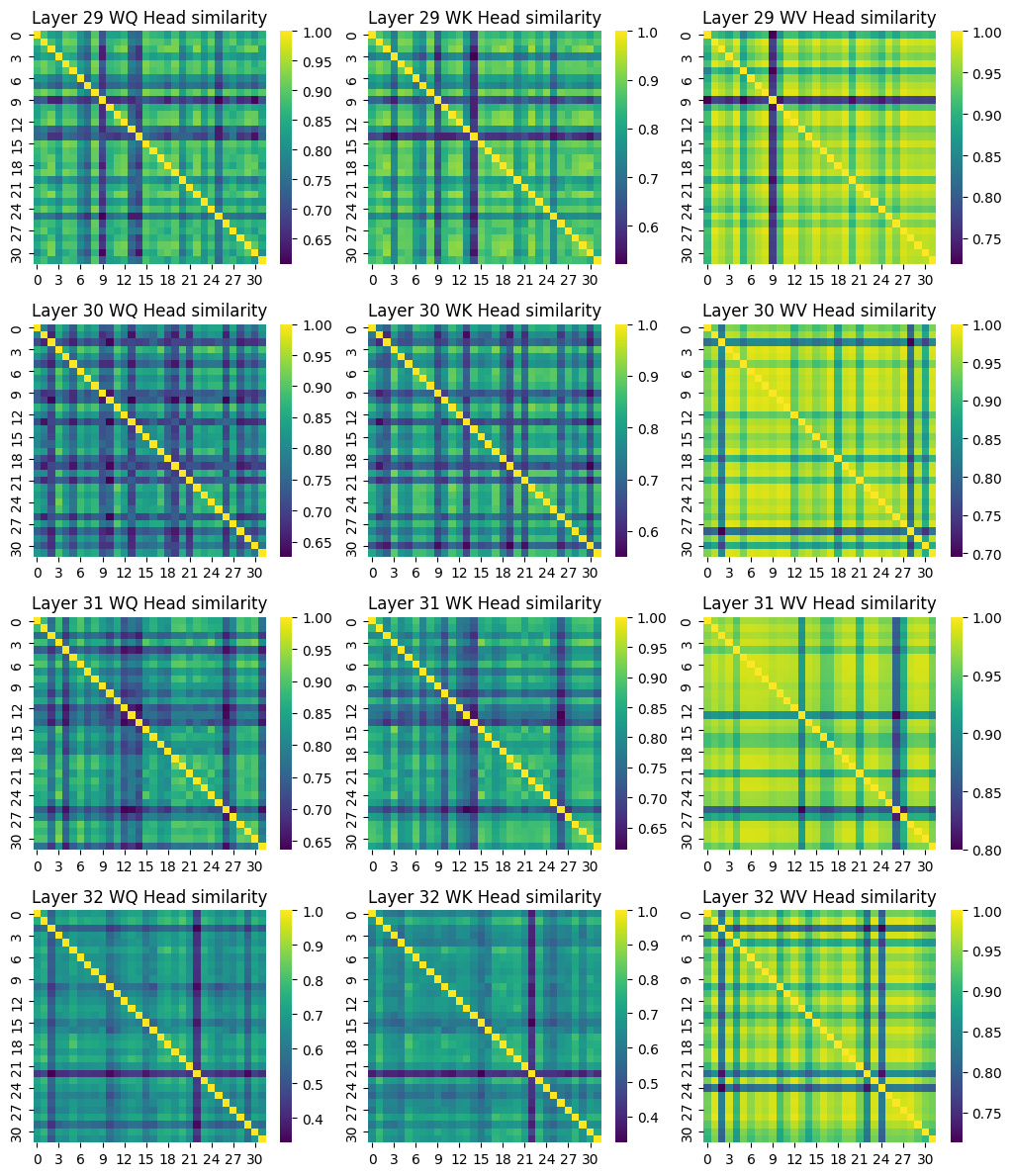}
        \end{subfigure}
    \end{tabular}
    \label{fig:2x2}
    \caption{Visualization of query, key, value head parameters similarity from layer 17 to layer 32 in LLaMA2-7B.}
\end{figure}

\newpage
\subsection{Header parameter characteristics in DHA}\label{app:obsdha}

The DHA parameter distribution of  shows consistency with MHA's. It indicates that DHA effectively aggregates multiple similar functional heads within clusters and new fused heads successfully reconstruct the functionalities of multiple origin heads in MHA. It is noteworthy that the significant reduction in the number of similar heads within the DHA architecture indicates that our method effectively reduces redundancy among the heads.

\begin{figure}[htbp]
    \centering
    \begin{tabular}{cc}
        \begin{subfigure}[b]{0.49\textwidth}
            \centering
            \includegraphics[width=\textwidth]{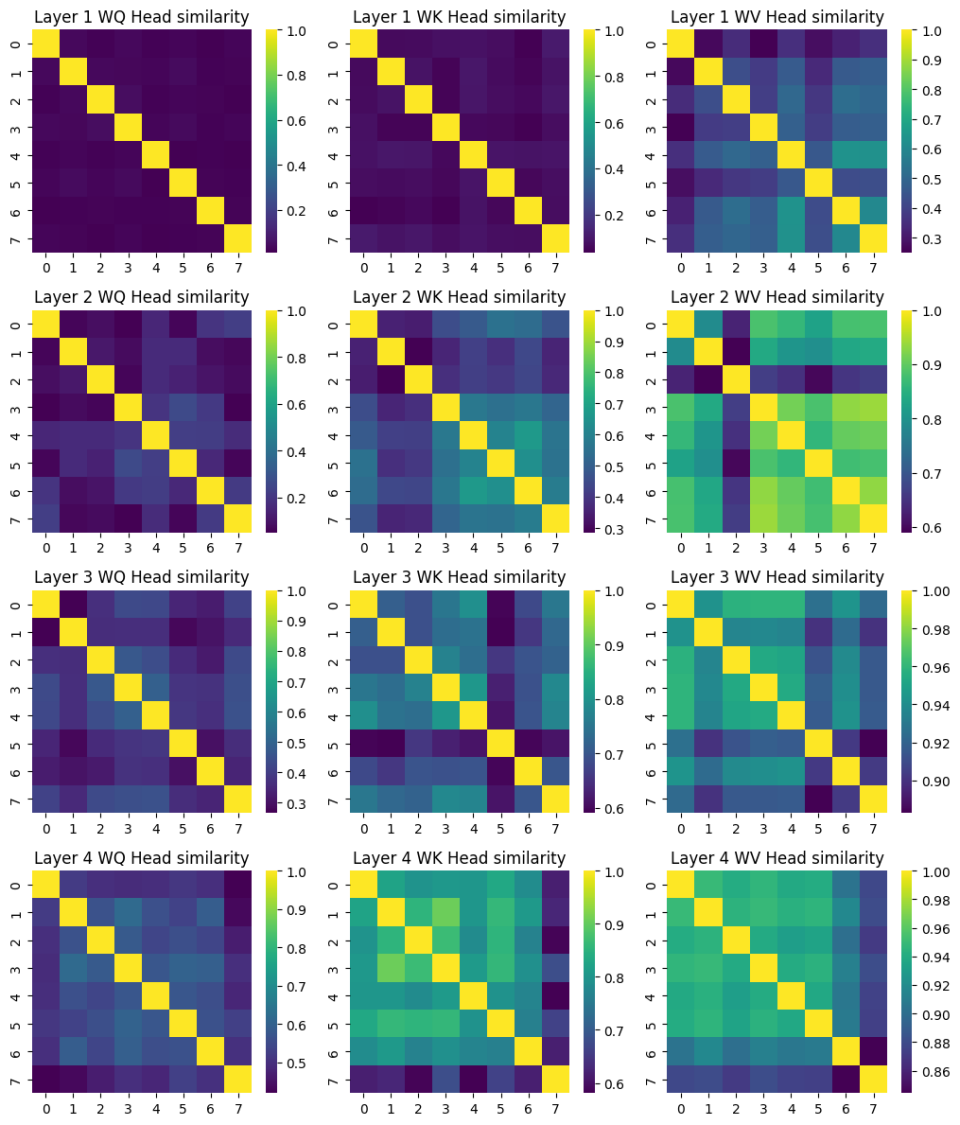}
        \end{subfigure} &
        \begin{subfigure}[b]{0.49\textwidth}
            \centering
            \includegraphics[width=\textwidth]{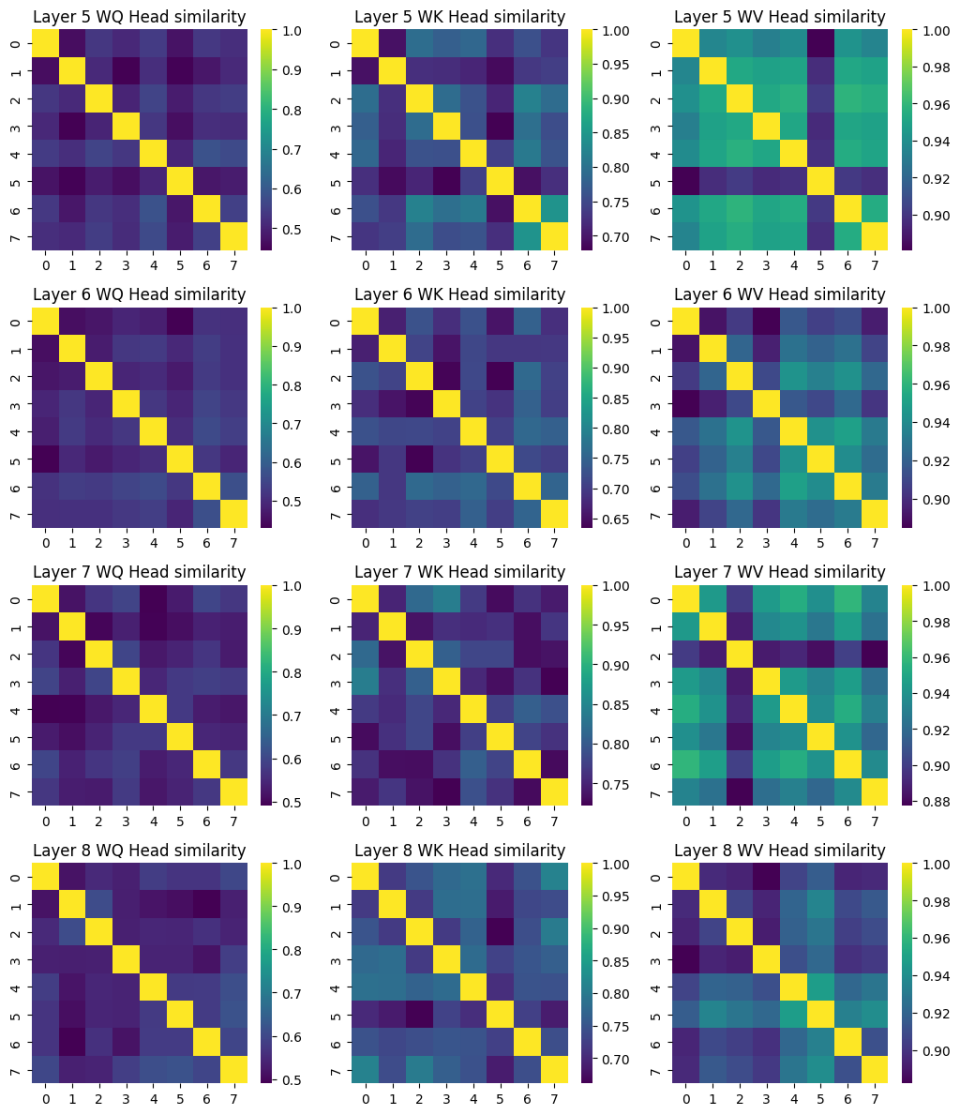}
        \end{subfigure} \\
    \end{tabular}
    \label{fig:2x2}
    \caption{Visualization of query, key, value head parameters similarity from layer 1 to layer 8 in DHA-7B-25\%.}
\end{figure}


%% file: checklist.tex
\newpage

\section*{NeurIPS Paper Checklist}

\begin{enumerate}

\item {\bf Claims}
    \item[] Question: Do the main claims made in the abstract and introduction accurately reflect the paper's contributions and scope?
    \item[] Answer: \answerYes{} 
    \item[] Justification:   The abstract provides a concise summary of the key findings and experiment results.  The introduction in Sec.~\ref{sec.intro} outlines the research questions and objectives in paragraph 3,4 and contribution in paragraph 5. 
    \item[] Guidelines:
    \begin{itemize}
        \item The answer NA means that the abstract and introduction do not include the claims made in the paper.
        \item The abstract and/or introduction should clearly state the claims made, including the contributions made in the paper and important assumptions and limitations. A No or NA answer to this question will not be perceived well by the reviewers. 
        \item The claims made should match theoretical and experimental results, and reflect how much the results can be expected to generalize to other settings. 
        \item It is fine to include aspirational goals as motivation as long as it is clear that these goals are not attained by the paper. 
    \end{itemize}

\item {\bf Limitations}
    \item[] Question: Does the paper discuss the limitations of the work performed by the authors?
    \item[] Answer: \answerYes{} 
    \item[] Justification: The paper discusses the limitations of the work performed by the authors in detail in Appendix Appendix.~\ref{app:limitation}, highlighting two specific limitations and the broader impact.
    \item[] Guidelines:
    \begin{itemize}
        \item The answer NA means that the paper has no limitation while the answer No means that the paper has limitations, but those are not discussed in the paper. 
        \item The authors are encouraged to create a separate "Limitations" section in their paper.
        \item The paper should point out any strong assumptions and how robust the results are to violations of these assumptions (e.g., independence assumptions, noiseless settings, model well-specification, asymptotic approximations only holding locally). The authors should reflect on how these assumptions might be violated in practice and what the implications would be.
        \item The authors should reflect on the scope of the claims made, e.g., if the approach was only tested on a few datasets or with a few runs. In general, empirical results often depend on implicit assumptions, which should be articulated.
        \item The authors should reflect on the factors that influence the performance of the approach. For example, a facial recognition algorithm may perform poorly when image resolution is low or images are taken in low lighting. Or a speech-to-text system might not be used reliably to provide closed captions for online lectures because it fails to handle technical jargon.
        \item The authors should discuss the computational efficiency of the proposed algorithms and how they scale with dataset size.
        \item If applicable, the authors should discuss possible limitations of their approach to address problems of privacy and fairness.
        \item While the authors might fear that complete honesty about limitations might be used by reviewers as grounds for rejection, a worse outcome might be that reviewers discover limitations that aren't acknowledged in the paper. The authors should use their best judgment and recognize that individual actions in favor of transparency play an important role in developing norms that preserve the integrity of the community. Reviewers will be specifically instructed to not penalize honesty concerning limitations.
    \end{itemize}

\item {\bf Theory Assumptions and Proofs}
    \item[] Question: For each theoretical result, does the paper provide the full set of assumptions and a complete (and correct) proof?
    \item[] Answer: \answerYes{} 
    \item[] Justification: This paper is mainly based on observation, making conjectures and methods and proving the effects through experiments. The paper defines the background in Sec.~\ref{sec.bg}, presents the conjecture in Sec.~\ref{sec.obs}, and provides a detailed derivation of the form and optimization process in Sec.~\ref{sec.method}. All assumptions made in the paper are thoroughly validated through experiments in Sec.~\ref{sec.exp}.
    \item[] Guidelines:
    \begin{itemize}
        \item The answer NA means that the paper does not include theoretical results. 
        \item All the theorems, formulas, and proofs in the paper should be numbered and cross-referenced.
        \item All assumptions should be clearly stated or referenced in the statement of any theorems.
        \item The proofs can either appear in the main paper or the supplemental material, but if they appear in the supplemental material, the authors are encouraged to provide a short proof sketch to provide intuition. 
        \item Inversely, any informal proof provided in the core of the paper should be complemented by formal proofs provided in appendix or supplemental material.
        \item Theorems and Lemmas that the proof relies upon should be properly referenced. 
    \end{itemize}

    \item {\bf Experimental Result Reproducibility}
    \item[] Question: Does the paper fully disclose all the information needed to reproduce the main experimental results of the paper to the extent that it affects the main claims and/or conclusions of the paper (regardless of whether the code and data are provided or not)?
    \item[] Answer:  \answerYes{} 
    \item[] Justification: The paper provides a detailed description of the experimental data setup and hyperparameter settings in Sec.~\ref{sec.exp}. Additionally, in Sec.~\ref{sec.method} and in Appendix.~\ref{app:implement} sections we thoroughly explain the derivation and implementation process, ensuring all necessary information for reproducing the main experimental results is disclosed.
    \item[] Guidelines:
    \begin{itemize}
        \item The answer NA means that the paper does not include experiments.
        \item If the paper includes experiments, a No answer to this question will not be perceived well by the reviewers: Making the paper reproducible is important, regardless of whether the code and data are provided or not.
        \item If the contribution is a dataset and/or model, the authors should describe the steps taken to make their results reproducible or verifiable. 
        \item Depending on the contribution, reproducibility can be accomplished in various ways. For example, if the contribution is a novel architecture, describing the architecture fully might suffice, or if the contribution is a specific model and empirical evaluation, it may be necessary to either make it possible for others to replicate the model with the same dataset, or provide access to the model. In general. releasing code and data is often one good way to accomplish this, but reproducibility can also be provided via detailed instructions for how to replicate the results, access to a hosted model (e.g., in the case of a large language model), releasing of a model checkpoint, or other means that are appropriate to the research performed.
        \item While NeurIPS does not require releasing code, the conference does require all submissions to provide some reasonable avenue for reproducibility, which may depend on the nature of the contribution. For example
        \begin{enumerate}
            \item If the contribution is primarily a new algorithm, the paper should make it clear how to reproduce that algorithm.
            \item If the contribution is primarily a new model architecture, the paper should describe the architecture clearly and fully.
            \item If the contribution is a new model (e.g., a large language model), then there should either be a way to access this model for reproducing the results or a way to reproduce the model (e.g., with an open-source dataset or instructions for how to construct the dataset).
            \item We recognize that reproducibility may be tricky in some cases, in which case authors are welcome to describe the particular way they provide for reproducibility. In the case of closed-source models, it may be that access to the model is limited in some way (e.g., to registered users), but it should be possible for other researchers to have some path to reproducing or verifying the results.
        \end{enumerate}
    \end{itemize}

\item {\bf Open access to data and code}
    \item[] Question: Does the paper provide open access to the data and code, with sufficient instructions to faithfully reproduce the main experimental results, as described in supplemental material?
    \item[] Answer: \answerNo{} 
    \item[] Justification: The datasets, baseline methods, and models used in the paper are fully open-source and available on Hugging Face. The paper includes the key implementation steps and code in Sec.~\ref{sec.method} and the Appendix.~\ref{app:implement}. However, the complete code is still being organized and is under consideration for open sourcing.
    \item[] Guidelines:
    \begin{itemize}
        \item The answer NA means that paper does not include experiments requiring code.
        \item Please see the NeurIPS code and data submission guidelines (\url{https://nips.cc/public/guides/CodeSubmissionPolicy}) for more details.
        \item While we encourage the release of code and data, we understand that this might not be possible, so “No” is an acceptable answer. Papers cannot be rejected simply for not including code, unless this is central to the contribution (e.g., for a new open-source benchmark).
        \item The instructions should contain the exact command and environment needed to run to reproduce the results. See the NeurIPS code and data submission guidelines (\url{https://nips.cc/public/guides/CodeSubmissionPolicy}) for more details.
        \item The authors should provide instructions on data access and preparation, including how to access the raw data, preprocessed data, intermediate data, and generated data, etc.
        \item The authors should provide scripts to reproduce all experimental results for the new proposed method and baselines. If only a subset of experiments are reproducible, they should state which ones are omitted from the script and why.
        \item At submission time, to preserve anonymity, the authors should release anonymized versions (if applicable).
        \item Providing as much information as possible in supplemental material (appended to the paper) is recommended, but including URLs to data and code is permitted.
    \end{itemize}

\item {\bf Experimental Setting/Details}
    \item[] Question: Does the paper specify all the training and test details (e.g., data splits, hyperparameters, how they were chosen, type of optimizer, etc.) necessary to understand the results?
    \item[] Answer: \answerYes{}, 
    \item[] Justification: The paper provides a detailed description of the experimental data setup and hyperparameter settings in Sec.~\ref{sec.exp}. Additionally, in Sec.~\ref{sec.method} and in Appendix.~\ref{app:implement} sections we thoroughly explain the derivation and implementation process.
    \item[] Guidelines:
    \begin{itemize}
        \item The answer NA means that the paper does not include experiments.
        \item The experimental setting should be presented in the core of the paper to a level of detail that is necessary to appreciate the results and make sense of them.
        \item The full details can be provided either with the code, in appendix, or as supplemental material.
    \end{itemize}

\item {\bf Experiment Statistical Significance}
    \item[] Question: Does the paper report error bars suitably and correctly defined or other appropriate information about the statistical significance of the experiments?
    \item[] Answer: \answerYes{} 
    \item[] Justification: All results are averaged over multiple tests, and we report the mean accuracy along with the standard deviation (acc\_norm) as a measure of error bars.
    \item[] Guidelines:
    \begin{itemize}
        \item The answer NA means that the paper does not include experiments.
        \item The authors should answer "Yes" if the results are accompanied by error bars, confidence intervals, or statistical significance tests, at least for the experiments that support the main claims of the paper.
        \item The factors of variability that the error bars are capturing should be clearly stated (for example, train/test split, initialization, random drawing of some parameter, or overall run with given experimental conditions).
        \item The method for calculating the error bars should be explained (closed form formula, call to a library function, bootstrap, etc.)
        \item The assumptions made should be given (e.g., Normally distributed errors).
        \item It should be clear whether the error bar is the standard deviation or the standard error of the mean.
        \item It is OK to report 1-sigma error bars, but one should state it. The authors should preferably report a 2-sigma error bar than state that they have a 96\% CI, if the hypothesis of Normality of errors is not verified.
        \item For asymmetric distributions, the authors should be careful not to show in tables or figures symmetric error bars that would yield results that are out of range (e.g. negative error rates).
        \item If error bars are reported in tables or plots, The authors should explain in the text how they were calculated and reference the corresponding figures or tables in the text.
    \end{itemize}

\item {\bf Experiments Compute Resources}
    \item[] Question: For each experiment, does the paper provide sufficient information on the computer resources (type of compute workers, memory, time of execution) needed to reproduce the experiments?
    \item[] Answer:  \answerYes{} 
    \item[] Justification: In Sec.~\ref{sec:exp.setup}, we report the GPUs we used, the memory, and detailed training information. For more information you can refer to the Appendix~\ref{app:implement}.
    \item[] Guidelines:
    \begin{itemize}
        \item The answer NA means that the paper does not include experiments.
        \item The paper should indicate the type of compute workers CPU or GPU, internal cluster, or cloud provider, including relevant memory and storage.
        \item The paper should provide the amount of compute required for each of the individual experimental runs as well as estimate the total compute. 
        \item The paper should disclose whether the full research project required more compute than the experiments reported in the paper (e.g., preliminary or failed experiments that didn't make it into the paper). 
    \end{itemize}
    
\item {\bf Code Of Ethics}
    \item[] Question: Does the research conducted in the paper conform, in every respect, with the NeurIPS Code of Ethics \url{https://neurips.cc/public/EthicsGuidelines}?
    \item[] Answer:\answerYes{} 
    \item[] Justification: The discussion of the ethics and impact can be consulted in Appendix.~\ref{app:limitation}. We are open and transparent throughout the study and do not design for human subjects, privacy data bias, or other issues.
    \item[] Guidelines:
    \begin{itemize}
        \item The answer NA means that the authors have not reviewed the NeurIPS Code of Ethics.
        \item If the authors answer No, they should explain the special circumstances that require a deviation from the Code of Ethics.
        \item The authors should make sure to preserve anonymity (e.g., if there is a special consideration due to laws or regulations in their jurisdiction).
    \end{itemize}

\item {\bf Broader Impacts}
    \item[] Question: Does the paper discuss both potential positive societal impacts and negative societal impacts of the work performed?
    \item[] Answer: \answerYes{} 
    \item[] Justification: The discussion of the broader impacts can be consulted in Appendix.~\ref{app:limitation}.
    \item[] Guidelines:
    \begin{itemize}
        \item The answer NA means that there is no societal impact of the work performed.
        \item If the authors answer NA or No, they should explain why their work has no societal impact or why the paper does not address societal impact.
        \item Examples of negative societal impacts include potential malicious or unintended uses (e.g., disinformation, generating fake profiles, surveillance), fairness considerations (e.g., deployment of technologies that could make decisions that unfairly impact specific groups), privacy considerations, and security considerations.
        \item The conference expects that many papers will be foundational research and not tied to particular applications, let alone deployments. However, if there is a direct path to any negative applications, the authors should point it out. For example, it is legitimate to point out that an improvement in the quality of generative models could be used to generate deepfakes for disinformation. On the other hand, it is not needed to point out that a generic algorithm for optimizing neural networks could enable people to train models that generate Deepfakes faster.
        \item The authors should consider possible harms that could arise when the technology is being used as intended and functioning correctly, harms that could arise when the technology is being used as intended but gives incorrect results, and harms following from (intentional or unintentional) misuse of the technology.
        \item If there are negative societal impacts, the authors could also discuss possible mitigation strategies (e.g., gated release of models, providing defenses in addition to attacks, mechanisms for monitoring misuse, mechanisms to monitor how a system learns from feedback over time, improving the efficiency and accessibility of ML).
    \end{itemize}
    
\item {\bf Safeguards}
    \item[] Question: Does the paper describe safeguards that have been put in place for responsible release of data or models that have a high risk for misuse (e.g., pretrained language models, image generators, or scraped datasets)?
    \item[] Answer:  \answerNA{} 
    \item[] Justification: This paper presents an improved approach based on the existing model architecture, but does not release any new models. The paper poses no such risks.
    \item[] Guidelines:
    \begin{itemize}
        \item The answer NA means that the paper poses no such risks.
        \item Released models that have a high risk for misuse or dual-use should be released with necessary safeguards to allow for controlled use of the model, for example by requiring that users adhere to usage guidelines or restrictions to access the model or implementing safety filters. 
        \item Datasets that have been scraped from the Internet could pose safety risks. The authors should describe how they avoided releasing unsafe images.
        \item We recognize that providing effective safeguards is challenging, and many papers do not require this, but we encourage authors to take this into account and make a best faith effort.
    \end{itemize}

\item {\bf Licenses for existing assets}
    \item[] Question: Are the creators or original owners of assets (e.g., code, data, models), used in the paper, properly credited and are the license and terms of use explicitly mentioned and properly respected?
    \item[] Answer: \answerYes{} 
    \item[] Justification: This article uses assets reasonably in compliance with the license, and the assets used are cited in the article.
    \item[] Guidelines:
    \begin{itemize}
        \item The answer NA means that the paper does not use existing assets.
        \item The authors should cite the original paper that produced the code package or dataset.
        \item The authors should state which version of the asset is used and, if possible, include a URL.
        \item The name of the license (e.g., CC-BY 4.0) should be included for each asset.
        \item For scraped data from a particular source (e.g., website), the copyright and terms of service of that source should be provided.
        \item If assets are released, the license, copyright information, and terms of use in the package should be provided. For popular datasets, \url{paperswithcode.com/datasets} has curated licenses for some datasets. Their licensing guide can help determine the license of a dataset.
        \item For existing datasets that are re-packaged, both the original license and the license of the derived asset (if it has changed) should be provided.
        \item If this information is not available online, the authors are encouraged to reach out to the asset's creators.
    \end{itemize}

\item {\bf New Assets}
    \item[] Question: Are new assets introduced in the paper well documented and is the documentation provided alongside the assets?
    \item[] Answer: \answerNA{} 
    \item[] Justification:  The paper does not release new assets.
    \item[] Guidelines:
    \begin{itemize}
        \item The answer NA means that the paper does not release new assets.
        \item Researchers should communicate the details of the dataset/code/model as part of their submissions via structured templates. This includes details about training, license, limitations, etc. 
        \item The paper should discuss whether and how consent was obtained from people whose asset is used.
        \item At submission time, remember to anonymize your assets (if applicable). You can either create an anonymized URL or include an anonymized zip file.
    \end{itemize}

\item {\bf Crowdsourcing and Research with Human Subjects}
    \item[] Question: For crowdsourcing experiments and research with human subjects, does the paper include the full text of instructions given to participants and screenshots, if applicable, as well as details about compensation (if any)? 
    \item[] Answer: \answerNA{} 
    \item[] Justification: The paper does not involve crowdsourcing nor research with human subjects
    \item[] Guidelines:
    \begin{itemize}
        \item The answer NA means that the paper does not involve crowdsourcing nor research with human subjects.
        \item Including this information in the supplemental material is fine, but if the main contribution of the paper involves human subjects, then as much detail as possible should be included in the main paper. 
        \item According to the NeurIPS Code of Ethics, workers involved in data collection, curation, or other labor should be paid at least the minimum wage in the country of the data collector. 
    \end{itemize}

\item {\bf Institutional Review Board (IRB) Approvals or Equivalent for Research with Human Subjects}
    \item[] Question: Does the paper describe potential risks incurred by study participants, whether such risks were disclosed to the subjects, and whether Institutional Review Board (IRB) approvals (or an equivalent approval/review based on the requirements of your country or institution) were obtained?
    \item[] Answer: \answerNA{} 
    \item[] Justification: The paper does not involve crowdsourcing nor research with human subjects.
    \item[] Guidelines:
    \begin{itemize}
        \item The answer NA means that the paper does not involve crowdsourcing nor research with human subjects.
        \item Depending on the country in which research is conducted, IRB approval (or equivalent) may be required for any human subjects research. If you obtained IRB approval, you should clearly state this in the paper. 
        \item We recognize that the procedures for this may vary significantly between institutions and locations, and we expect authors to adhere to the NeurIPS Code of Ethics and the guidelines for their institution. 
        \item For initial submissions, do not include any information that would break anonymity (if applicable), such as the institution conducting the review.
    \end{itemize}

\end{enumerate}

%% file: main.bbl
\begin{thebibliography}{10}

\bibitem{claude}
Anthropic.
\newblock Introducing claude.
\newblock 2023.

\bibitem{Gpt-4}
OpenAI.
\newblock Gpt-4 technical report.
\newblock {\em ArXiv}, page abs/2303.08774, 2023.

\bibitem{touvronLlamaOpenFoundation2023}
Hugo Touvron, Louis Martin, Kevin Stone, Peter Albert, Amjad Almahairi, Yasmine Babaei, Nikolay Bashlykov, Soumya Batra, Prajjwal Bhargava, Shruti Bhosale, Dan Bikel, Lukas Blecher, Cristian~Canton Ferrer, Moya Chen, Guillem Cucurull, David Esiobu, Jude Fernandes, Jeremy Fu, Wenyin Fu, Brian Fuller, Cynthia Gao, Vedanuj Goswami, Naman Goyal, Anthony Hartshorn, Saghar Hosseini, Rui Hou, Hakan Inan, Marcin Kardas, Viktor Kerkez, Madian Khabsa, Isabel Kloumann, Artem Korenev, Punit~Singh Koura, Marie-Anne Lachaux, Thibaut Lavril, Jenya Lee, Diana Liskovich, Yinghai Lu, Yuning Mao, Xavier Martinet, Todor Mihaylov, Pushkar Mishra, Igor Molybog, Yixin Nie, Andrew Poulton, Jeremy Reizenstein, Rashi Rungta, Kalyan Saladi, Alan Schelten, Ruan Silva, Eric~Michael Smith, Ranjan Subramanian, Xiaoqing~Ellen Tan, Binh Tang, Ross Taylor, Adina Williams, Jian~Xiang Kuan, Puxin Xu, Zheng Yan, Iliyan Zarov, Yuchen Zhang, Angela Fan, Melanie Kambadur, Sharan Narang, Aurelien Rodriguez, Robert Stojnic, Sergey Edunov, and Thomas
  Scialom.
\newblock Llama 2: {Open} {Foundation} and {Fine}-{Tuned} {Chat} {Models}, July 2023.

\bibitem{shazeer2019fast}
Noam Shazeer.
\newblock Fast transformer decoding: One write-head is all you need.
\newblock {\em arXiv preprint arXiv:1911.02150}, 2019.

\bibitem{ainslie2023gqa}
Joshua Ainslie, James Lee-Thorp, Michiel de~Jong, Yury Zemlyanskiy, Federico Lebr{\'o}n, and Sumit Sanghai.
\newblock Gqa: Training generalized multi-query transformer models from multi-head checkpoints.
\newblock {\em arXiv preprint arXiv:2305.13245}, 2023.

\bibitem{javadi2023gqkva}
Farnoosh Javadi, Walid Ahmed, Habib Hajimolahoseini, Foozhan Ataiefard, Mohammad Hassanpour, Saina Asani, Austin Wen, Omar~Mohamed Awad, Kangling Liu, and Yang Liu.
\newblock Gqkva: Efficient pre-training of transformers by grouping queries, keys, and values, 2023.

\bibitem{agarwal2024chai}
Saurabh Agarwal, Bilge Acun, Basil Homer, Mostafa Elhoushi, Yejin Lee, Shivaram Venkataraman, Dimitris Papailiopoulos, and Carole-Jean Wu.
\newblock Chai: Clustered head attention for efficient llm inference, 2024.

\bibitem{h2o}
Zhenyu Zhang, Ying Sheng, Tianyi Zhou, Tianlong Chen, Lianmin Zheng, Ruisi Cai, Zhao Song, Yuandong Tian, Christopher R{\'e}, Clark Barrett, et~al.
\newblock H $ \_2 $ o: Heavy-hitter oracle for efficient generative inference of large language models.
\newblock {\em arXiv preprint arXiv:2306.14048}, 2023.

\bibitem{liu2023scissorhands}
Zichang Liu, Aditya Desai, Fangshuo Liao, Weitao Wang, Victor Xie, Zhaozhuo Xu, Anastasios Kyrillidis, and Anshumali Shrivastava.
\newblock Scissorhands: Exploiting the persistence of importance hypothesis for llm kv cache compression at test time.
\newblock {\em arXiv preprint arXiv:2305.17118}, 2023.

\bibitem{ge2023model}
Suyu Ge, Yunan Zhang, Liyuan Liu, Minjia Zhang, Jiawei Han, and Jianfeng Gao.
\newblock Model tells you what to discard: Adaptive kv cache compression for llms.
\newblock {\em arXiv preprint arXiv:2310.01801}, 2023.

\bibitem{sink}
Guangxuan Xiao, Yuandong Tian, Beidi Chen, Song Han, and Mike Lewis.
\newblock Efficient streaming language models with attention sinks.
\newblock {\em arXiv preprint arXiv:2309.17453}, 2023.

\bibitem{qinExploringModeConnectivity2022}
Yujia Qin, Cheng Qian, Jing Yi, Weize Chen, Yankai Lin, Xu~Han, Zhiyuan Liu, Maosong Sun, and Jie Zhou.
\newblock Exploring {Mode} {Connectivity} for {Pre}-trained {Language} {Models}, October 2022.

\bibitem{zhangCRaShClusteringRemoving2023}
Kaiyan Zhang, Ning Ding, Biqing Qi, Xuekai Zhu, Xinwei Long, and Bowen Zhou.
\newblock {CRaSh}: {Clustering}, {Removing}, and {Sharing} {Enhance} {Fine}-tuning without {Full} {Large} {Language} {Model}, October 2023.

\bibitem{rush-etal-2010-dual}
Alexander~M. Rush, David Sontag, Michael Collins, and Tommi Jaakkola.
\newblock On dual decomposition and linear programming relaxations for natural language processing.
\newblock In Hang Li and Llu{\'\i}s M{\`a}rquez, editors, {\em Proceedings of the 2010 Conference on Empirical Methods in Natural Language Processing}, pages 1--11, Cambridge, MA, October 2010. Association for Computational Linguistics.

\bibitem{wangStructuredPruningLarge2020a}
Ziheng Wang, Jeremy Wohlwend, and Tao Lei.
\newblock Structured {Pruning} of {Large} {Language} {Models}.
\newblock In {\em Proceedings of the 2020 {Conference} on {Empirical} {Methods} in {Natural} {Language} {Processing} ({EMNLP})}, 2020.

\bibitem{xiaShearedLLaMAAccelerating2023}
Mengzhou Xia, Tianyu Gao, Zhiyuan Zeng, and Danqi Chen.
\newblock Sheared {LLaMA}: {Accelerating} {Language} {Model} {Pre}-training via {Structured} {Pruning}, October 2023.

\bibitem{vaswani2017attention}
Ashish Vaswani, Noam Shazeer, Niki Parmar, Jakob Uszkoreit, Llion Jones, Aidan~N Gomez, {\L}ukasz Kaiser, and Illia Polosukhin.
\newblock Attention is all you need.
\newblock {\em Advances in neural information processing systems}, 30, 2017.

\bibitem{kornblith2019similarity}
Simon Kornblith, Mohammad Norouzi, Honglak Lee, and Geoffrey Hinton.
\newblock Similarity of neural network representations revisited, 2019.

\bibitem{Redpajama}
TogetherAI.
\newblock Redpajama: An open source recipe to reproduce llama training dataset, 2023.

\bibitem{mosaicml2022composer}
The Mosaic~ML Team.
\newblock composer.
\newblock \url{https://github.com/mosaicml/composer/}, 2021.

\bibitem{eval-harness}
Leo Gao, Jonathan Tow, Baber Abbasi, Stella Biderman, Sid Black, Anthony DiPofi, Charles Foster, Laurence Golding, Jeffrey Hsu, Alain Le~Noac'h, Haonan Li, Kyle McDonell, Niklas Muennighoff, Chris Ociepa, Jason Phang, Laria Reynolds, Hailey Schoelkopf, Aviya Skowron, Lintang Sutawika, Eric Tang, Anish Thite, Ben Wang, Kevin Wang, and Andy Zou.
\newblock A framework for few-shot language model evaluation, 12 2023.

\bibitem{sciqa}
Johannes Welbl, Nelson~F. Liu, and Matt Gardner.
\newblock Crowdsourcing multiple choice science questions.
\newblock In Leon Derczynski, Wei Xu, Alan Ritter, and Tim Baldwin, editors, {\em Proceedings of the 3rd Workshop on Noisy User-generated Text, NUT@EMNLP 2017, Copenhagen, Denmark, September 7, 2017}, pages 94--106. Association for Computational Linguistics, 2017.

\bibitem{piqa}
Yonatan Bisk, Rowan Zellers, Ronan~Le Bras, Jianfeng Gao, and Yejin Choi.
\newblock {PIQA:} reasoning about physical commonsense in natural language.
\newblock In {\em The Thirty-Fourth {AAAI} Conference on Artificial Intelligence, {AAAI} 2020, The Thirty-Second Innovative Applications of Artificial Intelligence Conference, {IAAI} 2020, The Tenth {AAAI} Symposium on Educational Advances in Artificial Intelligence, {EAAI} 2020, New York, NY, USA, February 7-12, 2020}, pages 7432--7439. {AAAI} Press, 2020.

\bibitem{WinoGrande:conf/aaai/SakaguchiBBC20}
Keisuke Sakaguchi, Ronan~Le Bras, Chandra Bhagavatula, and Yejin Choi.
\newblock Winogrande: An adversarial winograd schema challenge at scale.
\newblock In {\em The Thirty-Fourth {AAAI} Conference on Artificial Intelligence, {AAAI} 2020, The Thirty-Second Innovative Applications of Artificial Intelligence Conference, {IAAI} 2020, The Tenth {AAAI} Symposium on Educational Advances in Artificial Intelligence, {EAAI} 2020, New York, NY, USA, February 7-12, 2020}, pages 8732--8740. {AAAI} Press, 2020.

\bibitem{clark2018think}
Peter Clark, Isaac Cowhey, Oren Etzioni, Tushar Khot, Ashish Sabharwal, Carissa Schoenick, and Oyvind Tafjord.
\newblock Think you have solved question answering? try arc, the ai2 reasoning challenge.
\newblock {\em arXiv preprint arXiv:1803.05457}, 2018.

\bibitem{HellaSwag:conf/acl/ZellersHBFC19}
Rowan Zellers, Ari Holtzman, Yonatan Bisk, Ali Farhadi, and Yejin Choi.
\newblock Hellaswag: Can a machine really finish your sentence?
\newblock In Anna Korhonen, David~R. Traum, and Llu{\'{\i}}s M{\`{a}}rquez, editors, {\em Proceedings of the 57th Conference of the Association for Computational Linguistics, {ACL} 2019, Florence, Italy, July 28- August 2, 2019, Volume 1: Long Papers}, pages 4791--4800. Association for Computational Linguistics, 2019.

\bibitem{arcChallenge:journals/corr/abs-1803-05457}
Peter Clark, Isaac Cowhey, Oren Etzioni, Tushar Khot, Ashish Sabharwal, Carissa Schoenick, and Oyvind Tafjord.
\newblock Think you have solved question answering? try arc, the {AI2} reasoning challenge.
\newblock {\em CoRR}, abs/1803.05457, 2018.

\bibitem{liu2020logiqa}
Jian Liu, Leyang Cui, Hanmeng Liu, Dandan Huang, Yile Wang, and Yue Zhang.
\newblock Logiqa: A challenge dataset for machine reading comprehension with logical reasoning.
\newblock {\em arXiv preprint arXiv:2007.08124}, 2020.

\bibitem{clark2019boolq}
Christopher Clark, Kenton Lee, Ming-Wei Chang, Tom Kwiatkowski, Michael Collins, and Kristina Toutanova.
\newblock Boolq: Exploring the surprising difficulty of natural yes/no questions.
\newblock {\em arXiv preprint arXiv:1905.10044}, 2019.

\bibitem{paperno2016lambada}
Denis Paperno, Germ{\'a}n Kruszewski, Angeliki Lazaridou, Quan~Ngoc Pham, Raffaella Bernardi, Sandro Pezzelle, Marco Baroni, Gemma Boleda, and Raquel Fern{\'a}ndez.
\newblock The lambada dataset: Word prediction requiring a broad discourse context.
\newblock {\em arXiv preprint arXiv:1606.06031}, 2016.

\bibitem{ouyang2022instructgpt}
Long Ouyang, Jeffrey Wu, Xu~Jiang, Diogo Almeida, Carroll Wainwright, Pamela Mishkin, Chong Zhang, Sandhini Agarwal, Katarina Slama, Alex Ray, et~al.
\newblock Training language models to follow instructions with human feedback.
\newblock {\em Advances in neural information processing systems}, 35, 2022.

\bibitem{alpaca}
Rohan Taori, Ishaan Gulrajani, Tianyi Zhang, Yann Dubois, Xuechen Li, Carlos Guestrin, Percy Liang, and Tatsunori~B. Hashimoto.
\newblock Stanford alpaca: An instruction-following llama model, 2023.

\bibitem{dubois2023alpacafarm}
Yann Dubois, Xuechen Li, Rohan Taori, Tianyi Zhang, Ishaan Gulrajani, Jimmy Ba, Carlos Guestrin, Percy Liang, and Tatsunori~B Hashimoto.
\newblock Alpacafarm: A simulation framework for methods that learn from human feedback.
\newblock {\em arXiv preprint arXiv:2305.14387}, 2023.

\bibitem{DBLP:conf/cvpr/XieXP17}
Di~Xie, Jiang Xiong, and Shiliang Pu.
\newblock All you need is beyond a good init: Exploring better solution for training extremely deep convolutional neural networks with orthonormality and modulation.
\newblock In {\em 2017 {IEEE} Conference on Computer Vision and Pattern Recognition, {CVPR} 2017, Honolulu, HI, USA, July 21-26, 2017}, pages 5075--5084. {IEEE} Computer Society, 2017.

\bibitem{DBLP:conf/nips/WuW019}
Lemeng Wu, Dilin Wang, and Qiang Liu.
\newblock Splitting steepest descent for growing neural architectures.
\newblock In Hanna~M. Wallach, Hugo Larochelle, Alina Beygelzimer, Florence d'Alch{\'{e}}{-}Buc, Emily~B. Fox, and Roman Garnett, editors, {\em Advances in Neural Information Processing Systems 32: Annual Conference on Neural Information Processing Systems 2019, NeurIPS 2019, December 8-14, 2019, Vancouver, BC, Canada}, pages 10655--10665, 2019.

\bibitem{chen2015net2net}
Tianqi Chen, Ian Goodfellow, and Jonathon Shlens.
\newblock Net2net: Accelerating learning via knowledge transfer.
\newblock {\em arXiv preprint arXiv:1511.05641}, 2015.

\bibitem{wangLearningGrowPretrained2023}
Peihao Wang, Rameswar Panda, Lucas~Torroba Hennigen, Philip Greengard, Leonid Karlinsky, Rogerio Feris, David~Daniel Cox, Zhangyang Wang, and Yoon Kim.
\newblock Learning to {Grow} {Pretrained} {Models} for {Efficient} {Transformer} {Training}, March 2023.

\bibitem{tay2020efficient}
Yi~Tay, Mostafa Dehghani, Dara Bahri, and Donald Metzler.
\newblock Efficient transformers: A survey.
\newblock {\em arXiv preprint arXiv:2009.06732}, 2020.

\bibitem{sparse}
Rewon Child, Scott Gray, Alec Radford, and Ilya Sutskever.
\newblock Generating long sequences with sparse transformers.
\newblock {\em arXiv preprint arXiv:1904.10509}, 2019.

\bibitem{kitaev2020reformer}
Nikita Kitaev, Lukasz Kaiser, and Anselm Levskaya.
\newblock Reformer: The efficient transformer.
\newblock In {\em 8th International Conference on Learning Representations, {ICLR} 2020, Addis Ababa, Ethiopia, April 26-30, 2020}. OpenReview.net, 2020.

\bibitem{bigbird}
Manzil Zaheer, Guru Guruganesh, Avinava Dubey, Joshua Ainslie, Chris Alberti, Santiago Ontanon, Philip Pham, Anirudh Ravula, Qifan Wang, Li~Yang, et~al.
\newblock Big bird: Transformers for longer sequences.
\newblock {\em Advances in neural information processing systems}, 2020.

\bibitem{beltagy2020longformer}
Iz~Beltagy, Matthew~E Peters, and Arman Cohan.
\newblock Longformer: The long-document transformer.
\newblock {\em arXiv preprint arXiv:2004.05150}, 2020.

\bibitem{transformer_xl}
Zihang Dai, Zhilin Yang, Yiming Yang, Jaime~G. Carbonell, Quoc~V. Le, and Ruslan Salakhutdinov.
\newblock Transformer-xl: Attentive language models beyond a fixed-length context.
\newblock {\em CoRR}, abs/1901.02860, 2019.

\bibitem{erniedoc}
Siyu Ding, Junyuan Shang, Shuohuan Wang, Yu~Sun, Hao Tian, Hua Wu, and Haifeng Wang.
\newblock Ernie-doc: A retrospective long-document modeling transformer.
\newblock {\em arXiv preprint arXiv:2012.15688}, 2020.

\bibitem{rmt}
Aydar Bulatov, Yury Kuratov, and Mikhail Burtsev.
\newblock Recurrent memory transformer.
\newblock {\em Advances in Neural Information Processing Systems}, 35:11079--11091, 2022.

\bibitem{chevalier2023adapting}
Alexis Chevalier, Alexander Wettig, Anirudh Ajith, and Danqi Chen.
\newblock Adapting language models to compress contexts.
\newblock {\em arXiv preprint arXiv:2305.14788}, 2023.

\bibitem{msrnn}
Matanel Oren, Michael Hassid, Yossi Adi, and Roy Schwartz.
\newblock Transformers are multi-state rnns.
\newblock {\em arXiv preprint arXiv:2401.06104}, 2024.

\bibitem{chen-etal-2024-nacl}
Yilong Chen, Guoxia Wang, Junyuan Shang, Shiyao Cui, Zhenyu Zhang, Tingwen Liu, Shuohuan Wang, Yu~Sun, Dianhai Yu, and Hua Wu.
\newblock {NACL}: A general and effective {KV} cache eviction framework for {LLM} at inference time.
\newblock In Lun-Wei Ku, Andre Martins, and Vivek Srikumar, editors, {\em Proceedings of the 62nd Annual Meeting of the Association for Computational Linguistics (Volume 1: Long Papers)}, pages 7913--7926, Bangkok, Thailand, August 2024. Association for Computational Linguistics.

\bibitem{dettmers2022llmint8}
Tim Dettmers, Mike Lewis, Younes Belkada, and Luke Zettlemoyer.
\newblock Llm.int8(): 8-bit matrix multiplication for transformers at scale, 2022.

\bibitem{frantar2023gptq}
Elias Frantar, Saleh Ashkboos, Torsten Hoefler, and Dan Alistarh.
\newblock Gptq: Accurate post-training quantization for generative pre-trained transformers, 2023.

\bibitem{gray1998quantization}
Robert~M. Gray and David~L. Neuhoff.
\newblock Quantization.
\newblock {\em IEEE transactions on information theory}, 44(6):2325--2383, 1998.

\bibitem{hinton2015distilling}
Geoffrey Hinton, Oriol Vinyals, and Jeff Dean.
\newblock Distilling the knowledge in a neural network, 2015.

\bibitem{Gou_2021}
Jianping Gou, Baosheng Yu, Stephen~J. Maybank, and Dacheng Tao.
\newblock Knowledge distillation: A survey.
\newblock {\em International Journal of Computer Vision}, 129(6):1789–1819, March 2021.

\bibitem{cai2022pile}
Lianshang Cai, Linhao Zhang, Dehong Ma, Jun Fan, Daiting Shi, Yi~Wu, Zhicong Cheng, Simiu Gu, and Dawei Yin.
\newblock Pile: Pairwise iterative logits ensemble for multi-teacher labeled distillation.
\newblock In {\em Proceedings of the 2022 Conference on Empirical Methods in Natural Language Processing: Industry Track}, pages 587--595, 2022.

\bibitem{pmlr-v202-liu23am}
Zichang Liu, Jue Wang, Tri Dao, Tianyi Zhou, Binhang Yuan, Zhao Song, Anshumali Shrivastava, Ce~Zhang, Yuandong Tian, Christopher Re, and Beidi Chen.
\newblock Deja vu: Contextual sparsity for efficient {LLM}s at inference time.
\newblock In Andreas Krause, Emma Brunskill, Kyunghyun Cho, Barbara Engelhardt, Sivan Sabato, and Jonathan Scarlett, editors, {\em Proceedings of the 40th International Conference on Machine Learning}, volume 202 of {\em Proceedings of Machine Learning Research}, pages 22137--22176. PMLR, 23--29 Jul 2023.

\bibitem{frantarSparseGPTMassiveLanguage2023}
Elias Frantar and Dan Alistarh.
\newblock {SparseGPT}: {Massive} {Language} {Models} {Can} {Be} {Accurately} {Pruned} in {One}-{Shot}, March 2023.

\bibitem{zhang2020accelerating}
Minjia Zhang and Yuxiong He.
\newblock Accelerating training of transformer-based language models with progressive layer dropping.
\newblock {\em Advances in Neural Information Processing Systems}, 33:14011--14023, 2020.

\bibitem{sajjad2023effect}
Hassan Sajjad, Fahim Dalvi, Nadir Durrani, and Preslav Nakov.
\newblock On the effect of dropping layers of pre-trained transformer models.
\newblock {\em Computer Speech \& Language}, 77:101429, 2023.

\bibitem{dao2022flashattention}
Tri Dao, Dan Fu, Stefano Ermon, Atri Rudra, and Christopher R{\'e}.
\newblock Flashattention: Fast and memory-efficient exact attention with io-awareness.
\newblock {\em Advances in Neural Information Processing Systems}, 35:16344--16359, 2022.

\bibitem{240102415LLaMA}
[2401.02415] {LLaMA} {Pro}: {Progressive} {LLaMA} with {Block} {Expansion}.

\bibitem{chen-etal-2024-lemon}
Yilong Chen, Junyuan Shang, Zhenyu Zhang, Shiyao Cui, Tingwen Liu, Shuohuan Wang, Yu~Sun, and Hua Wu.
\newblock {LEMON}: Reviving stronger and smaller {LM}s from larger {LM}s with linear parameter fusion.
\newblock In Lun-Wei Ku, Andre Martins, and Vivek Srikumar, editors, {\em Proceedings of the 62nd Annual Meeting of the Association for Computational Linguistics (Volume 1: Long Papers)}, pages 8005--8019, Bangkok, Thailand, August 2024. Association for Computational Linguistics.

\end{thebibliography}
